\documentclass[11pt]{article}

\usepackage{acl}

\usepackage{times}
\usepackage{latexsym}

\usepackage{algorithm}
\usepackage{algpseudocode}

\usepackage{booktabs}
\usepackage{makecell}
\usepackage{multirow}

\usepackage[T1]{fontenc}

\usepackage[utf8]{inputenc}

\usepackage{microtype}
\usepackage{amssymb}
\usepackage{amsmath}

\usepackage{inconsolata}

\usepackage{graphicx}

\title{TENP: Trapezoidal Expert Neuron Pruning For Mixture-of-Experts}

\author{
 \textbf{Jiangyang He},
 \textbf{Shaolin Zhu},
 \textbf{Deyi Xiong\footnotemark[1]}
\\
 TJUNLP Lab, School of Computer Science and Technology, Tianjin University, China
\\
 \texttt{\{jiangyanghe, zhushaolin, dyxiong\}@tju.edu.cn}
}

\begin{document}
\maketitle

\renewcommand{\thefootnote}{\fnsymbol{footnote}}
\footnotetext[1]{Corresponding author.}

\begin{abstract}
Mixture-of-Experts large language models (LLMs) scale efficiently through sparse activation, yet their deployment is fundamentally constrained by the large static parameter footprint of experts. Existing compression approaches either remove entire experts, disrupting routing topology and harming performance, or rely on unstructured weight pruning with limited practical efficiency. To address the limitations, we propose TENP, a structured \textbf{T}rapezoidal \textbf{E}xpert \textbf{N}euron \textbf{P}runing framework. Using a few samples, we identify and retain important experts, while applying expert neuron pruning (ENP) to less important experts, preserving model parameters in a trapezoidal pattern from shallow to deep layers. When evaluating expert importance, we jointly consider both the magnitude of the expert output and its ability to change the direction of the input vector. For ENP, we measure each neuron’s projected contribution to the expert output to identify and retain important neurons. We conduct extensive experiments on the Qwen and DeepSeek models. Under a routing expert sparsity of 40\% and an average of 63.76\% activated expert parameters, the DeepSeek model suffers only a 1-point drop in accuracy compared to the full-parameter model. Moreover, it outperforms the full-parameter model by 10\% on code generation tasks.
\end{abstract}

\section{Introduction}

The paradigm shift towards Mixture-of-Experts (MoE) architectures has become a cornerstone for scaling, as it effectively decouples model capacity from computational cost \cite{GPT-OSS}. By conditionally activating a sparse subset of parameters for each token, models such as DeepSeek-V3.2 \cite{DeepseekV3.2} and Qwen3 \cite{Qwen3} achieve state-of-the-art performance with significantly reduced FLOPs compared to their dense counterparts \cite{fuxitranyu, TibetanLLM, MultilingualLLM}. However, this efficiency comes with a substantial trade-off that the massive static parameter footprint required to host the full set of experts creates a severe bottleneck for deployment, particularly in memory-constrained environments \cite{DiEP}. 

To address the memory bottleneck, existing compression methodologies primarily bifurcate into expert pruning, weight pruning \cite{DiEP} or quantization \cite{QuantizationJiangcun, QuantizationRenRen}. Expert pruning methods attempt to permanently remove less significant experts based on activation frequency or router gradients \cite{SEER-MoE, DomainPruning}. However, these approaches are constrained by a fundamental limitation: the coarse-grained removal of entire experts disrupts the model's original routing topology. As indicated in recent analyses \cite{EAC-MoE}, altering the routing path compels forces tokens to be dispatched to sub-optimal experts, which can precipitate significant performance degradation on domain-specific tasks. Alternatively, unstructured weight pruning methods \cite{Wanda} target individual parameters but consequently yield irregular sparsity patterns that require specialized hardware kernels for acceleration.

Recent empirical analyses indicate that for compressed models, restoring their original routing paths can recover model performance \cite{EAC-MoE}. However, experts that have been removed can no longer be routed to. We therefore prune redundant parameters within experts while keeping the experts intact and routable. Prior work \cite{Wanda, NeuronExperts} has found that a large amount of redundancy exists at the microscopic, neuron-level within these experts. Furthermore, redundancy is non-uniformly distributed in the layer-wise representational capacity of LLMs. Shallow layers primarily process local syntactic features and exhibit high redundancy, whereas deep layers encapsulate complex semantic reasoning \cite{MoE-I2, MOLA}. This implies that a uniform pruning ratio is sub-optimal for MoE architectures and motivates a depth-aware allocation of parameter budgets.

To address these limitations, we propose \textbf{TENP} (Trapezoidal Expert Neuron Pruning), a structured pruning framework tailored for MoE LLMs. Unlike expert-level pruning, which alters where a token goes, TENP focuses on slimming down what the expert computes by pruning neurons within experts to maintain the validity of the pre-trained router's decisions. In particular, TENP introduces a depth-aware Trapezoidal sparsity strategy.
It applies aggressive pruning to shallow, high-redundancy layers and progressively retains more capacity in deep layers to preserve reasoning capabilities. To accurately identify redundant neurons without costly retraining, we design a dual-metric evaluation that combines the magnitude contribution with a directional impact score to distinguish between essential transformations and redundant identity-like mappings.

Our contributions are summarized as follows:

\begin{itemize}
\item We propose \textbf{TENP}, a \textbf{structured expert-neuron pruning} that reduces memory usage while strictly preserving the original MoE routing topology.
\item We introducean Trapezoidal sparsity distribution strategy. We empirically demonstrate that allocating high parameter budgets to deep layers while aggressively compressing shallow layers yields a superior trade-off between model size and performance.
\item Experiments demonstrate that, under routing expert sparsities of 40\% and 70\%, our method consistently outperforms existing expert pruning approaches across a wide range of reasoning tasks and benchmarks, achieving an average improvement of approximately 16\%.
\end{itemize}

\section{Related Work}

\paragraph{MoE LLMs}
MoE models have gained significant attention in recent years due to their unique capability of expanding model capacity without proportionally increasing computational costs. MoE architectures partition a large neural network (or specific components) into multiple expert sub-networks, where only a subset is activated for each input token based on routing decisions \cite{FirstMoE, SwitchTransformer}. GShard \cite{GShard} pioneers trillion-parameter models by distributing parameters across multiple devices. DeepSeekMoE \cite{DeepSeekMoE} presents a shared expert mechanism to reduce communication overhead and computational cost. The effectiveness of MoE architectures has been validated at the 16 billion parameter scale \cite{Qwen1.5MoE-A2.7B, DeepSeekV2Lite}. More recently, Mixtral \cite{Mixtral8x7B}, GPT-OSS \cite{GPT-OSS}, DeepSeekV3.2 \cite{DeepseekV3.2}, and KimiK2 \cite{KimiK2} have demonstrated MoE's efficacy at the hundred-billion parameter scale. Advanced routing strategies have also emerged, with DA-MoE \cite{DA-MoE} and XMoE \cite{XMoE} implementing dynamic expert selection mechanisms that allocate more computational resources to challenging tokens. \citet{MOLA} propose a pyramid-shaped architecture where layers closer to the output employ more parameters. Meanwhile, GroveMoE \cite{GroveMoE} adopts heterogeneous experts with dynamic parameter activation to optimize performance.

\begin{figure*}[t]
  \centering
  \includegraphics[width=\textwidth]{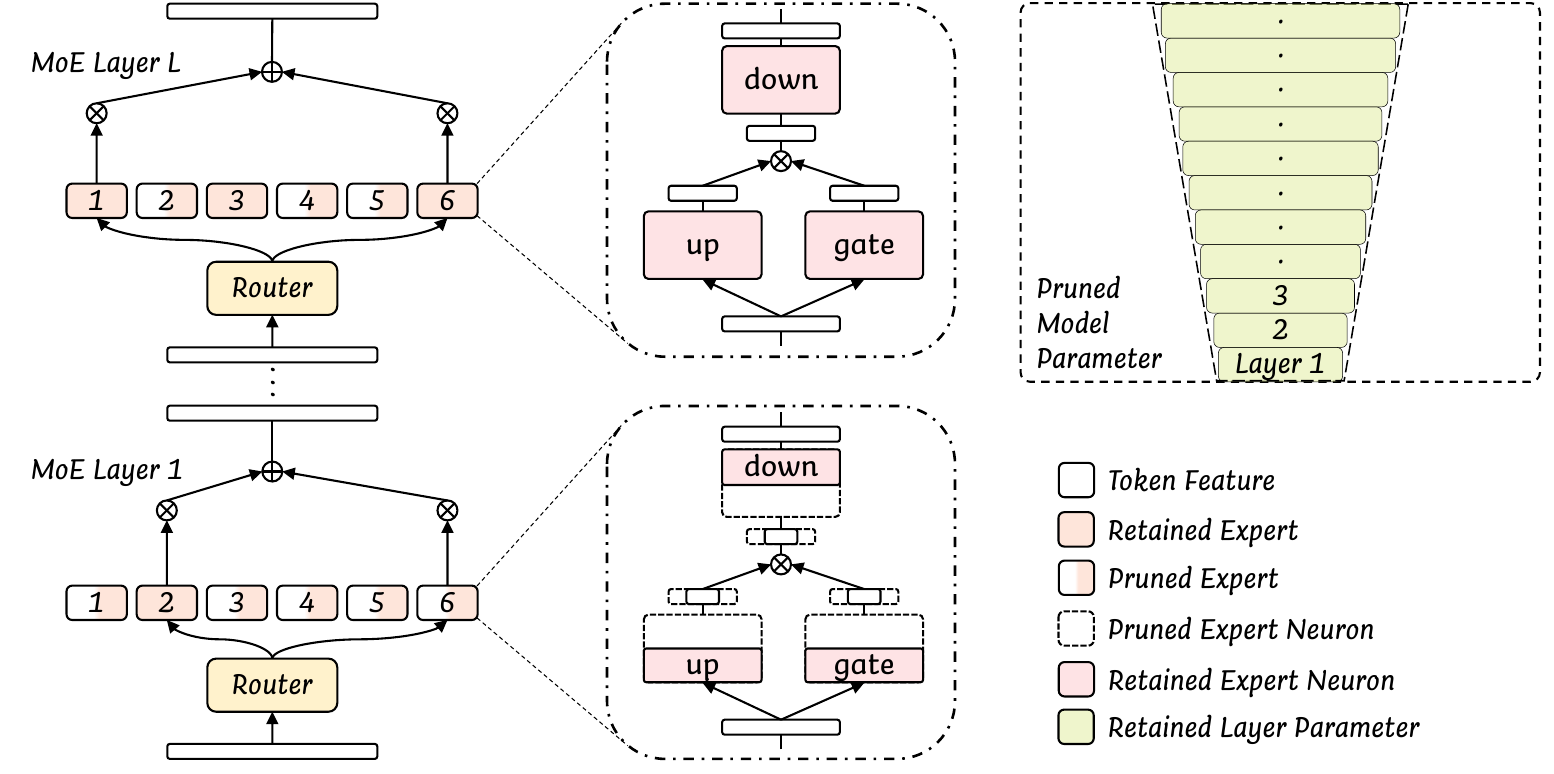}
  \caption{Schematic illustration of the Trapezoidal Expert Neuron Pruning (TENP) framework.
(a) TENP preserves the complete routing structure, allowing even pruned experts to remain routable.
(b) Comparison between an expert before pruning and after pruning: neurons inside the expert are removed, while the dimensionality of the expert’s output remains unchanged.
(c) After pruning, the parameter distribution of the model changes from a rectangular structure, where each layer has an identical number of parameters, to a trapezoidal structure with fewer parameters in shallow layers and more parameters in deep layers.}
  \label{fig:framework}
\end{figure*}

\paragraph{Pruning and Compression of MoE Models}
As scaling laws continue to drive exponential growth in MoE model sizes, numerous techniques have emerged to reduce parameter counts while preserving performance. SEER-MoE \cite{SEER-MoE} prunes less important experts based on their activation frequency or gating score. \citet{NotAllExpertsAreEqual} introduces expert-level pruning combined with dynamic skipping mechanisms. \citet{DomainPruning} concentrates model capabilities in specific domains through few-shot expert localization. While these methods reduce parameter counts, they generally fail to decrease computational requirements proportionally. \citet{NeuronExperts} achieve computation reduction through fine-grained neuron activation within selected experts, without reducing the overall parameter count. More comprehensive approaches include MoE-I2 \cite{MoE-I2}, which proposes a three-stage pruning methodology requiring subsequent fine-tuning to recover performance, and Task-Specific Expert Pruning \cite{TaskPurning}, which integrates pruning with task-specific training. SparseGPT \cite{SparseGPT}, MoE-Pruner \cite{MoE-Pruner}, and Wanda \cite{Wanda} employ an unstructured pruning method that imposes specific hardware requirements. Alternative approaches like \cite{NeuronMerge} consolidate important neurons across experts, though this compromises the original routing mechanism. Similarly, \citet{MergeThenCompress} and \citet{MergeExpert} propose expert merging strategies that face routing challenges. The methods compress all experts into a single dense model, although they have reduced many parameters, have changed the architecture of the model \cite{Merge2One,2Dense}.

\section{Method}

Our approach consists of two stages. In the first stage, we follow and modify EASY-EP to identify important experts. We conduct experiments with different important expert retention ratios, as detailed in Appendix~\ref{sec:ExpertRetentionRatios}.
Ultimately, we determine that under a routed expert sparsity of 40\%, retaining 20\%--30\% of important experts yields optimal results, while under a routed expert sparsity of 70\%, retaining 5\% of important experts is sufficient.
This approach not only preserves the complete routing topology but also reduces the average number of activated parameters per expert. Moreover, the number of important experts preserved in each layer gradually increases with depth. As illustrated in Figure \ref{fig:framework}, higher layers retain more parameters, resulting in a trapezoidal parameter distribution.
In the second stage, we perform neuron pruning on the less important experts identified in the first stage. Alternatively, the first stage can be skipped, and neuron pruning can be directly applied to all experts; we refer to this approach as Expert Neuron Pruning (ENP).
We evaluate the contribution of each intermediate dimension of an expert to the output using the method described below, where each intermediate dimension corresponds to specific rows and columns of the expert’s parameters. For dimensions deemed unimportant, we remove the corresponding rows and columns of parameters to eliminate that intermediate dimension.

\subsection{Retaining Important Experts}
To evaluate expert importance, we compute an importance score from the experts' outputs. Suppose the input to the MoE block at layer $l$ is $\mathbf{h}_t^l$. Let the output of the $i$-th routed expert $\mathrm{E}_i^l$ for token $t$ be $\mathbf{\overline{h}}_{i,t}^l$, and the routing weight be $\mathbf{g}_{i,t}^l$. The MoE output of all routed experts at layer $l$ is denoted as $\mathbf{\tilde{h}}_t^l$, which is the weighted sum of the $N$ experts' outputs, as follows:
\begin{equation}
\mathbf{\overline{h}}_{i,t}^l = \mathrm{E}_i^l(\mathbf{h}_t^l),
\end{equation}
\begin{equation}
\mathbf{\tilde{h}}_t^l = \sum_{i=1}^{N} \mathbf{g}_{i,t}^l \cdot \mathbf{\overline{h}}_{i,t}^l.
\end{equation}

We define the length-based contribution of expert outputs as $\mathbf{c}_{i,t}^l$. Using $\|\cdot\|$ to denote the $\ell_2$ norm, for each token $t$, we quantify the contribution of expert $i$ to the layer output by the product of the routing weight and the output norm:
\begin{equation}
\mathbf{c}_{i,t}^l = \mathbf{g}_{i,t}^l \, \| \mathbf{\overline{h}}_{i,t}^l \|,\quad \forall\, \mathbf{g}_{i,t}^l>0.
\end{equation}

Inspired by \citet{ShortGPT} and \citet{DomainPruning}, beyond the output magnitude and routing weight, we consider each expert’s ability to alter the direction of the input vector. \citet{ShortGPT} suggest that when a layer behaves closer to an identity mapping, it tends to be more redundant. However, $\mathbf{c}_{i,t}^l$ only reflects the magnitude of an expert’s output vector and the weight assigned to the expert by the router; it does not capture the expert’s ability to change the direction of the input vector. In other words, relying solely on $\mathbf{c}_{i,t}^l$ does not allow us to determine whether an expert is performing an identity mapping, since an expert that implements an identity mapping can also have a large $\mathbf{c}_{i,t}^l$ value. Therefore, we introduce $\mathbf{s}_{i,t}^l$ to quantify the directional change induced by the expert, defined as one minus the cosine similarity between the expert’s input and output. Values close to zero correspond to near-identity behavior, whereas larger values indicate more substantial angular deviations.

\begin{equation}
\mathbf{s}_{i,t}^l = 1 - \mathrm{Sim}(\mathbf{h}_{t}^l, \mathbf{\overline{h}}_{i,t}^l),
\end{equation}
where $\mathrm{Sim}(\cdot,\cdot)$ denotes cosine similarity.

Finally, we jointly consider the magnitude of the expert output vector, the weight assigned by the router, and the expert’s ability to alter the vector direction, and average these factors over all tokens $T$ to comprehensively evaluate the importance of each expert, as shown in the following formula:
\begin{equation}
\mathbf{I}(\mathrm{E}_i^l) = \sum_{t=1}^{T} \mathbf{c}_{i,t}^l \cdot \mathbf{s}_{i,t}^l.
\end{equation}

For each domain, we use 128 validation samples to evaluate expert and neuron importance. We also investigate the effect of different numbers of samples on the results, as reported in \ref{sec:data}. For aggregation across domains $\tau$, we apply an $\ell_2$-norm-based normalization (regularization) as follows:
\begin{equation}
\mathbf{I}_{\mathrm{mix}}(\mathrm{E}_i^l) =
\sum_{\tau \in \mathcal{T}}
\frac{\mathbf{I}_{\tau}(\mathrm{E}_i^l)}
{\sqrt{\sum_{j=1}^{N} \mathbf{I}_{\tau}(\mathrm{E}_j^l)^2}}.
\end{equation}


\subsection{Expert Neuron Pruning}

After selecting important experts, we perform neuron pruning on the remaining experts. As noted above, neuron pruning can also be applied to all experts directly. For a single expert, its output can be written as:
\begin{equation}
\overline{\mathbf{h}}_{t}^l = \mathbf{W}_{\mathrm{down}}\,
\mathrm{SwiGLU}(\mathbf{W}_{\mathrm{up}} \mathbf{h}_t^l,\; \mathbf{W}_{\mathrm{gate}} \mathbf{h}_t^l),
\end{equation}
where $\mathbf{W}_{\mathrm{gate}}$ is the gating matrix, $\mathbf{W}_{\mathrm{up}}$ is the first linear projection, $\mathbf{W}_{\mathrm{down}}$ is the second linear projection, and $\mathrm{SwiGLU}(\cdot,\cdot)$ denotes the SiLU-gated activation.\footnote{If using a different FFN variant, the formulation can be adjusted accordingly.}

We extract the $k$-th row of $\mathbf{W}_{\mathrm{up}}$ and $\mathbf{W}_{\mathrm{gate}}$, and the $k$-th column of $\mathbf{W}_{\mathrm{down}}$, denoted by $\mathbf{w}_{\mathrm{up},k}$, $\mathbf{w}_{\mathrm{gate},k}$, and $\mathbf{w}_{\mathrm{down},k}$, respectively. Substituting them into the above equation yields the expert output when only the $k$-th neuron is retained:
\begin{equation}
\overline{\mathbf{h}}_{t,k}^l =
\mathbf{w}_{\mathrm{down},k}\,
\mathrm{SwiGLU}(\mathbf{w}_{\mathrm{up},k} \mathbf{h}_t^l,\; \mathbf{w}_{\mathrm{gate},k} \mathbf{h}_t^l).
\end{equation}

Both $\overline{\mathbf{h}}_{t,k}^l$ and $\overline{\mathbf{h}}_{t}^l$ share the same output dimension, i.e.,
$\overline{\mathbf{h}}_{t,k}^l,\; \overline{\mathbf{h}}_{t}^l \in \mathbb{R}^{d}$.
We then quantify the importance of neuron $k$ by either the magnitude of its projection onto the full expert output or by the $\ell_2$ norm of $\overline{\mathbf{h}}_{t,k}^l$. Using the projection magnitude, we define:
\begin{equation}
\mathbf{p}_{k} =
\frac{\langle \overline{\mathbf{h}}_{t,k}^l,\; \overline{\mathbf{h}}_{t}^l \rangle}
{\| \overline{\mathbf{h}}_{t}^l \|}.
\end{equation}

A larger projection magnitude (or $\ell_2$ norm) indicates a more important neuron. We aggregate neuron importance by averaging across tokens for the $k$-th neuron in expert $i$ at layer $l$ (As in Appendix in Section \ref{sec:Algorithm}):
\begin{equation}
\mathbf{P}_{i,k}^l = \frac{1}{T}\sum_{t=1}^{T} \mathbf{P}_{i,k}^l.
\end{equation}

We prune each expert by keeping its top-$K$ most important neurons. Let $\mathrm{TopK}(\mathbf{P}_i^l)_{\mathrm{idx}}$ denote the indices of the top-$K$ values in $\mathbf{P}_i^l$. For the $i$-th expert at layer $l$, the pruned parameters are:

\begin{table*}[t]
\centering
\resizebox{\textwidth}{!}{
\begin{tabular}{ll*{8}{c}}
\toprule
Model &Method &E$\downarrow$ &A$\downarrow$ &GSM8K &MBPP &Humaneval &ARC-E &ARC-C &Avg.\\
\midrule
\multirow{13}*{\makecell{Qwen1.5MoE\\-A2.7B}}& Full & 100\% & 100\% &61.5 &47.6&34.2&86.4&76.1&61.16 \\
\cmidrule(lr){2-10}
& Random & 30\% & 100\% & 1.7 & 0.0  & 0.0  & 25.8 & 25.7  & 10.64 \\ 
& Frequency & 30\% & 100\% & 1.6 & 0.0  & 0.0  & 52.6  & 44.1  &19.66 \\ 
& Gating Score & 30\% & 100\% & 2.3  & 0.4 & 1.2 & 56.3  & 43.2   &20.68 \\ 
& EASY-EP & 30\% & 100\% & 3.4 & 0.4 & 1.8 & 55.3 & 44.8  &21.14 \\ 
& ENP(Ours) &30\% & \textbf{30\%} & 23.1 & 25.1 & 17.1 & 78.2 & 66.0   &41.90 \\ 
& TENP(Ours) & 30\% & 34.51\% & \textbf{25.7} & \textbf{25.3} & \textbf{18.9} & \textbf{78.2} & \textbf{66.9}  &\textbf{43.00} \\ 
\cmidrule(lr){2-10}
& Random &60\%&100\%&19.6&1.2&0.0&77.8&64.5&32.62 \\
& Frequency & 60\%&100\%&30.9&14.6&6.7&80.0&66.7&39.78 \\
& Gating Score & 60\%&100\%&30.8&18.9&8.5&84.5&73.0&43.14 \\
& EASY-EP & 60\%&100\%&36.8&35.4&19.5&80.1&68.3&48.02 \\
& ENP(Ours) &60\%&\textbf{60\%} &51.3 &40.6 &28.0  &84.6 &74.9 &55.88 \\
& TENP(Ours) & 60\%&61.38\%&\textbf{58.3}  & \textbf{45.7}  & \textbf{31.1} & \textbf{85.1}  & \textbf{75.1}  & \textbf{59.06} \\
\midrule
\multirow{13}*{\makecell{DeepSeek\\-V2-Lite}}& Full & 100\% & 100\% &41.1&\underline{43.2}&\underline{26.2}&84.1&70.3&52.98 \\
\cmidrule(lr){2-10}
& Random & 30\% & 100\% & 1.1 & 0.0  & 0.0  & 24.5 & 24.2  &9.96 \\ 
& Frequency & 30\% & 100\% & 1.9 & 0.0  & 0.0  & 24.3 & 24.0   &10.04 \\ 
& Gating Score & 30\% & 100\% & 1.9 & 0.0  & 0.0  & 26.7  & 27.8  &11.28 \\ 
& EASY-EP & 30\% & 100\% & 2.8 & 4.3 & 1.2  & 38.4 & 29.6  &15.26 \\ 
& ENP(Ours) & 30\% & \textbf{30\%} & 3.7 & 9.1 & 0.0  & 59.6 & 48.4  &24.16 \\ 
& TENP(Ours) & 30\% & 36.38\% & \textbf{21.1} & \textbf{33.1} & \textbf{14.0} & \textbf{73.3} & \textbf{57.1}  &\textbf{39.72} \\ 
\cmidrule(lr){2-10}
& Random &60\%&100\%&1.8&0.0&0.0&34.3&31.7&13.56 \\
& Frequency & 60\%&100\%&32.8&24.4&11.6&75.3&61.8&41.18 \\
& Gating Score & 60\%&100\%&21.5&30.7&11.6&74.0&57.5&39.06 \\
& EASY-EP & 60\%&100\%&34.8&42.1&14.6&76.7&62.7&46.18 \\
& ENP(Ours) & 60\%&\textbf{60\%}&22.1 &33.9 &19.5 &77.8 &65.7 &43.80 \\
& TENP(Ours) & 60\%&63.76\%&\textbf{38.4}&\underline{\textbf{45.7}}&\underline{\textbf{29.9}}&\textbf{79.1}
&\textbf{66.8}&\textbf{51.98} \\
\bottomrule
\end{tabular}
}
\caption{Comparison of our method with other expert-pruning approaches across all benchmarks. $E$ denotes the equivalent total parameter count of the routed experts, and $A$ denotes the average activated parameter count of the routed experts.}
\label{tab:MainResult}
\end{table*}

\begin{align}
\tilde{\mathbf{W}}_{i,\mathrm{up}}^l
&= \mathbf{W}_{i,\mathrm{up}}^l\big[\mathrm{TopK}(\mathbf{P}_i^l)_{\mathrm{idx}},:\big], \\
\tilde{\mathbf{W}}_{i,\mathrm{gate}}^l
&= \mathbf{W}_{i,\mathrm{gate}}^l\big[\mathrm{TopK}(\mathbf{P}_i^l)_{\mathrm{idx}},:\big], \\
\tilde{\mathbf{W}}_{i,\mathrm{down}}^l
&= \mathbf{W}_{i,\mathrm{down}}^l\big[:,\mathrm{TopK}(\mathbf{P}_i^l)_{\mathrm{idx}}\big].
\end{align}

By replacing the original expert’s weight matrix $\mathbf{W}_{i,\mathrm{up}}^l,\mathbf{W}_{i,\mathrm{gate}}^l,\mathbf{W}_{i,\mathrm{down}}^l$ with the pruned $\tilde{\mathbf{W}}_{i,\mathrm{up}}^l,\tilde{\mathbf{W}}_{i,\mathrm{gate}}^l\tilde{\mathbf{W}}_{i,\mathrm{down}}^l$, we obtain the neuron-pruned expert. The forward computation of the pruned expert can be formally formulated as:

\begin{equation}
\overline{\mathbf{h}}_{t}^l = \tilde{\mathbf{W}}_{\mathrm{down}}\,
\mathrm{SwiGLU}(\tilde{\mathbf{W}}_{\mathrm{up}} \mathbf{h}_t^l,\; \tilde{\mathbf{W}}_{\mathrm{gate}} \mathbf{h}_t^l),
\end{equation}

Neuron pruning provides an additional benefit: it not only reduces the total number of parameters, but also decreases the number of parameters that are activated in the routed experts.

\section{Experiment}

\begin{table*}[t]
\centering
\resizebox{\textwidth}{!}{
\begin{tabular}{l*{7}{c}}
\toprule
Method &E$\downarrow$ &A$\downarrow$ &MBPP &Humaneval &ARC-E &ARC-C &Avg.\\
\midrule
Full & 100\% & 100\% &43.2& 26.2& 84.2& 70.3& 55.98 \\
\cmidrule(lr){2-8}
Random &60\%&100\%&0.0&0.0&34.3&31.7&16.50 \\
TENP w/o Both (Random Select Both) &60\% &60.73\% &1.5 &0.0  &41.0  &37.0  &19.88\\
TENP w/o ENP (Random Select ENP) &60\% &60.87\% &6.7 &0.6 &59.8 &45.7 &28.20 \\
TENP w/o TE (Random Select TE) &60\% &61.00\% &34.7 &20.7 &76.8 &64.4 &49.15 \\
Only EP & 60\%&100\%&45.7 &8.5 &77.8 &63.6 &48.90 \\
Only ENP-L2 &60\%&\textbf{60\%}&32.3 &18.9 &77.5 &65.9 &48.65 \\
Only ENP-COS &60\%&\textbf{60\%}&41.7 &17.7 &78.7 &65.9 &51.00 \\
TENP & 60\%&63.55\%&\textbf{47.2}&\textbf{29.8}&\textbf{79.4}&\textbf{66.7}&\textbf{55.78} \\
\bottomrule
\end{tabular}
}
\caption{Results of the ablation study. Only EP denotes applying expert pruning only. ENP-L2 and ENP-COS denote neuron pruning based on neuron importance measured by the $\ell_2$ norm or by the projection length, respectively. Random denotes randomly selected groups used as a control baseline.}
\label{tab:AblationResult}
\end{table*}

We conducted experiments on two MoE models with distinct architectures: Qwen1.5-MoE-A2.7B \cite{Qwen1.5MoE-A2.7B} and DeepSeek-V2-Lite \cite{DeepSeekV2Lite}. Detailed model descriptions are provided in Appendix~\ref{sce:ModelDetail}. We also report experiments on Qwen3-Next-80B-A3B-Instruct\cite{Qwen3} in Appendix \ref{sec:80BModel}.

\subsection{Experimental Setup}

\paragraph{Evaluation.}
We evaluated the proposed method on a diverse set of benchmarks spanning multiple domains.
For challenging mathematical reasoning, we used GSM8K \cite{GSM8K} with 8-shot prompting.
For code generation, we reported results on MBPP \cite{MBPP} with 3-shot prompting and HumanEval \cite{HumanEval} under zero-shot evaluation.
For science question answering, we used ARC-Easy and ARC-Challenge \cite{ARC}, both under 25-shot prompting.
Following standard protocols, we used dataset-specific few-shot settings on open-source datasets and reported results averaged over three runs.

\paragraph{Baselines.}
We compared against four representative expert-pruning approaches for MoE models.
As a lower bound, we included a random expert selection baseline to quantify performance when no preference is given to expert selection.
We further evaluated the frequency-based pruning and gating-score-based pruning strategies proposed in SEER-MoE \cite{SEER-MoE}.
In addition, we compared with the recently proposed EASY-EP \cite{DomainPruning}.
These baselines retained only the experts that are ranked highest according to statistics estimated from a small number of samples per dataset.\footnote{We do not included comparisons to methods that require unstructured pruning (e.g., Wanda), additional fine-tuning (e.g., MoE-I2), or approaches that modify the model architecture.}

\subsection{Main Results}

Table \ref{tab:MainResult} reported a comprehensive comparison between our approach and a variety of baselines across multiple datasets, model backbones, and pruning ratios. When we applied TENP to prune DeepSeek-V2-Lite, the resulting model achieves better performance on both mathematical reasoning and knowledge-intensive QA tasks than other expert-pruning methods such as SEER-MoE and EASY-EP. On code-generation benchmarks (e.g., MBPP and HumanEval), the pruned model yields more than one point improvement over the full model. Similar trends are observed on Qwen1.5MoE-A2.7B, where our method consistently surpassed competing approaches across all evaluated domains. 
For relatively simple QA-style tasks such as ARC, pruned models generally preserved their original performance well. In contrast, on more challenging reasoning-heavy tasks (e.g., GSM8K), pruning induced some degradation; nevertheless, our approach still outperforms baselines. Notably, for code generation, the performance drop is often small and can even exceed the full model, as highlighted by the underlined entries in Table \ref{tab:MainResult}. We also conducted experiments at a higher sparsity level (70\%), where our method consistently outperformed other approaches across all benchmarks. Additional comparisons under different sparsity settings are provided in Appendix~\ref{sec:DiffSparsity}.

\paragraph{Activated-Parameter Efficiency.}
Most existing expert-pruning methods can remove experts, yet keep the number of experts selected by the router unchanged. As a result, although the total parameter count decreases, the number of activated parameters remained the same (normalized to $100\%$). In contrast, our approach pruned neurons within experts, thereby reducing not only the number of experts but also the activated parameters of routed experts accordingly. With neuron pruning alone, our ENP variant achieved the lowest activated-parameter footprint and, in most cases, still outperforms prior expert-pruning methods.

As indicated by metric $A$ in Table \ref{tab:MainResult}, if the router selected a preserved (unpruned) expert, its activated parameters remain at $100\%$. If it selected a neuron-pruned expert, the activated parameters are reduced to $50\%$ or lower of the original. Since the routing probability of preserved experts is higher than that of neuron-pruned experts, the overall activated parameters are only slightly higher than the fraction of preserved experts. Consequently, the activated-parameter cost of our routed experts is substantially lower than that of existing expert-pruning approaches.

\begin{figure*}[t]
  \centering
  \includegraphics[width=\textwidth]{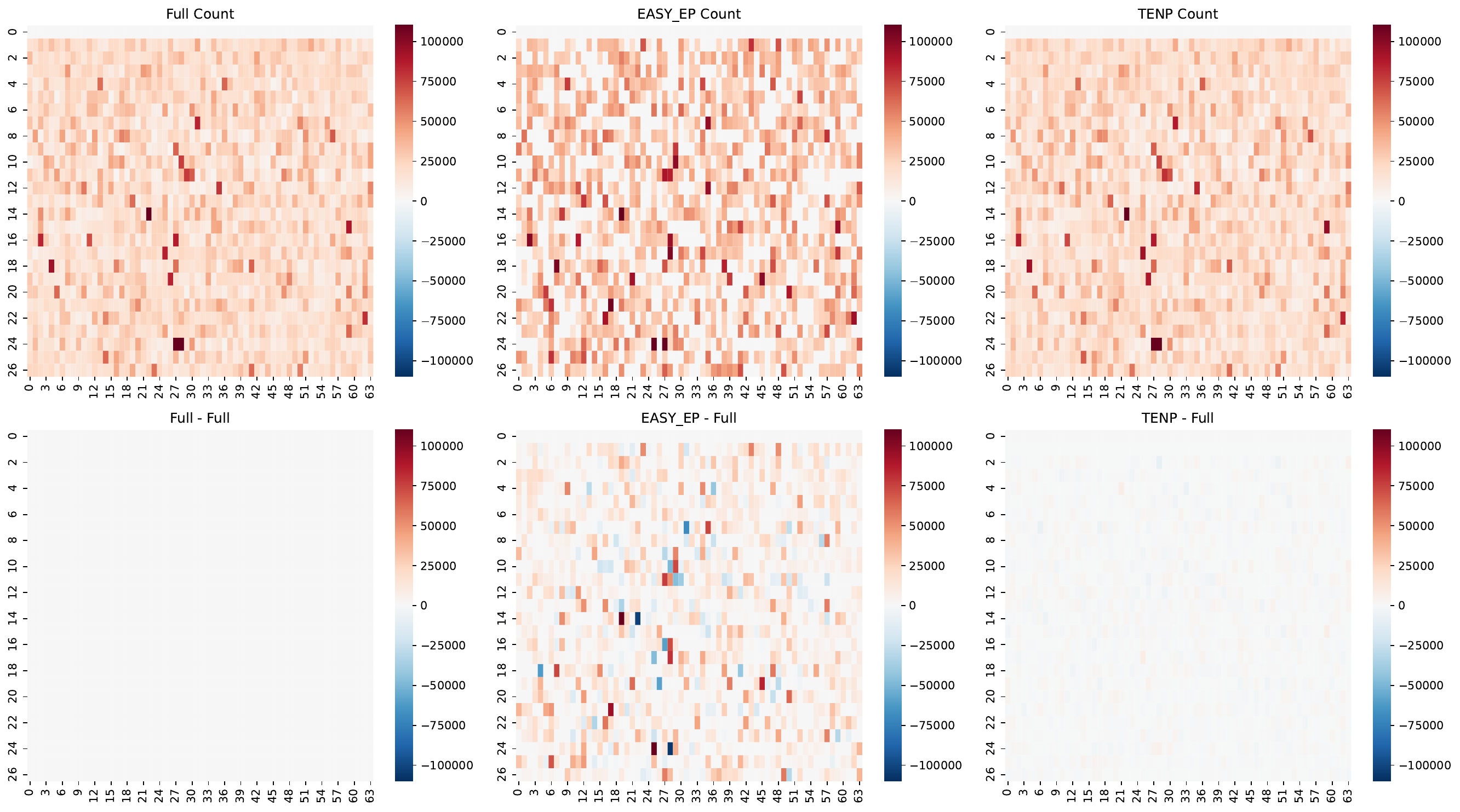}
  \caption{Expert selection frequencies under different pruning methods, and their differences compared to the expert selection frequencies of the full-parameter model.}
  \label{fig:expert_usage}
\end{figure*}

\begin{figure}[t]
\centering
  \includegraphics[width=\columnwidth]{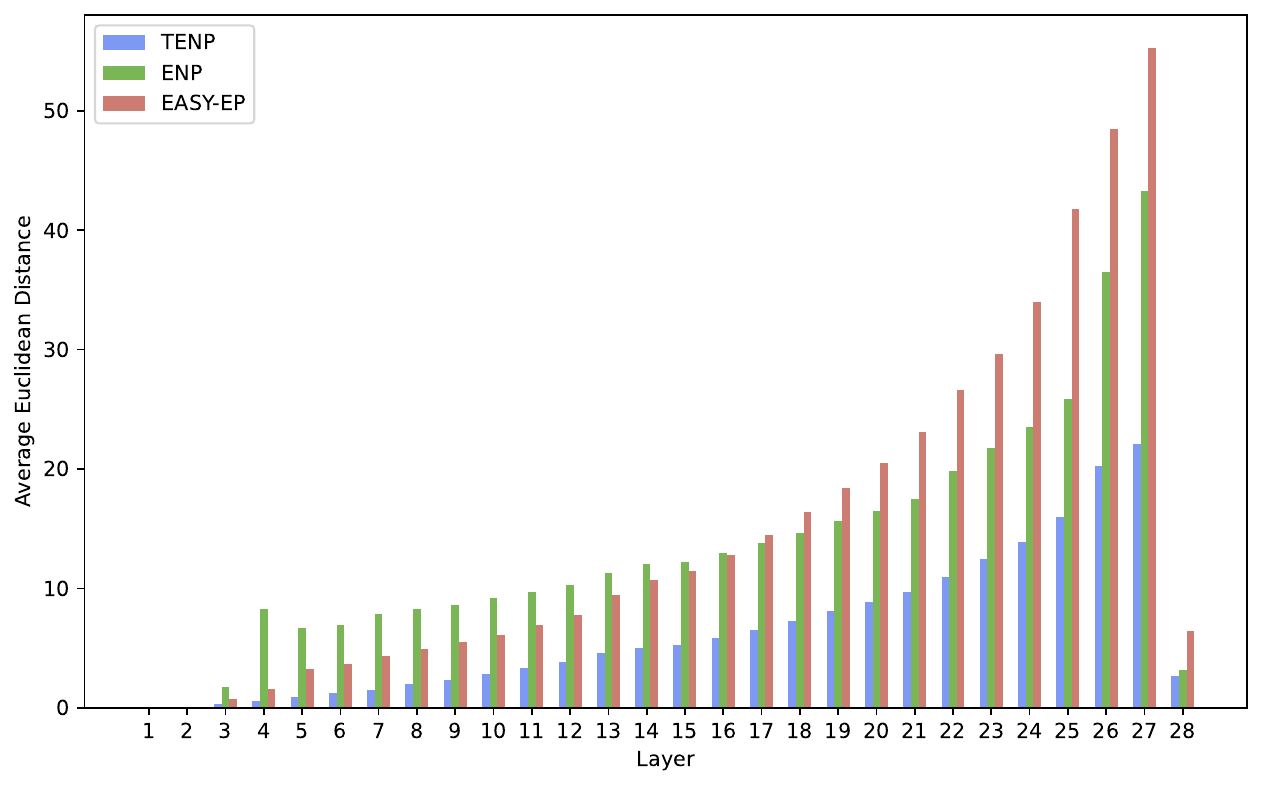}
  \caption{Layer-wise error of the pruned models, measured as the Euclidean distance between the output vectors of each layer and those of the full-parameter model.}
  \label{fig:diff_comparison}
\end{figure}

\subsection{Ablation Study}
\label{sec:ablation}

As shown in Table \ref{tab:AblationResult}, applying either expert pruning alone (Only EP) or expert neuron pruning alone (ENP) already yields a certain level of pruning effectiveness to validate the effectiveness of the two core components of our method.

When ENP is applied, measuring neuron importance by the projection length of a neuron's output vector onto the final output vector of the corresponding expert consistently outperformed the method that used the $\ell_2$ norm of the neuron output vector. This result indicated that projection-based importance better captured a neuron's contribution to the expert output. When experts are selected randomly, the model performance is noticeably affected. This observation demonstrated the validity of the experts we have retained in a trapezoidal structure. Furthermore, when both experts and neurons are selected randomly, our method still outperformed the baseline that randomly selected experts (Random), even though both configurations have the same number of parameters. This advantage arose from the different parameter distributions: our method allocated more parameters to higher layers through the trapezoidal structure and preserved the complete routing architecture. We also compare other layer selection methods in Appendix \ref{sec:OtherLayerSelectionMethods}. By combining important experts with neurons selected based on projection-length importance, we obtained TENP, which achieved the best overall performance. Notably, TENP incurred almost no accuracy loss compared to the full-parameter model.

\begin{table*}[t]
\centering
\resizebox{\textwidth}{!}{
\begin{tabular}{ll*{9}{c}}
\toprule
Method &Data &GSM8K &MBPP &Humaneval &ARC-E &ARC-C &InAvg. &OutAvg &InAvg. &OutAvg\\
\midrule
\multirow{3}*{EASY-EP}& Math &38.3 &28.0 &13.4 &64.4 &55.6 &38.30 &\textbf{40.35} &\multirow{3}*{47.45} &\multirow{3}*{31.81} \\
&Code &33.0 &43.7 &18.9 &56.3 &44.2 &31.30 &44.50 && \\
&Science &23.6 &6.3 &1.8 &79.1 &66.4 &72.75  &10.57 && \\
\midrule
\multirow{3}*{TENP}& Math &40.9 &20.5  &10.4  &66.8 &51.2 &\textbf{40.90}  &37.23 &\multirow{3}*{\textbf{51.37}} &\multirow{3}*{\textbf{33.06}} \\
&Code &29.5 &50.8 &27.4 &69.9 &54.1 &\textbf{39.10}  &\textbf{51.17} && \\
&Science &20.2 &9.8 &2.4 &80.7 &67.5 &\textbf{74.10}  &\textbf{10.80} && \\
\bottomrule
\end{tabular}
}
\caption{Generalization performance of the pruned models. Data denotes the domain of the data used for pruning.}
\label{tab:generalizationResult}
\end{table*}

\begin{table}[t]
\centering
\resizebox{0.95\linewidth}{!}{
    \begin{tabular}{llcc}
    \toprule
        Model & Method & E & MMLU  \\ 
    \midrule
        \multirow{4}*{\makecell{Qwen1.5MoE\\-A2.7B}} & Full & 100\% & 61.05  \\ 
    \cmidrule(lr){2-4}
        ~ & GatingScore & 60\% & 49.13  \\ 
        ~ & EASY-EP & 60\% & 47.49  \\ 
        ~ & TENP(Ours) & 60\% & \textbf{54.81}  \\ 
    \bottomrule
\end{tabular}
}
\caption{The generalization performance of three pruning methods on the MMLU dataset, where no MMLU data is used during the pruning process.}
\label{tab:AblationMMLU}
\end{table}

\subsection{Routing Analysis}
\label{sec:routing_analysis}
We characterized routing changes by measuring the difference in expert selection counts before and after pruning. As shown in Figure \ref{fig:expert_usage}, we analyzed and compared the expert selection statistics of the full-parameter model, the expert-pruned model, and our proposed pruning method on DeepSeek. We focus on the changes in expert selection frequencies induced by pruning. When we retained only a subset of experts, it induced substantial changes in expert selection patterns. Notably, the selection frequency of remaining experts does not uniformly increase after pruning. Instead, while some experts experience a significant increase in selection frequency, others are selected less frequently, or even less than before pruning. This observation indicates that routing behavior changes drastically.
More importantly, some experts that were not pruned and were previously routable are no longer selected by the router after pruning. This led to a significant shift in the output representations. Furthermore, we observed a trend that the magnitude of changes in expert selection frequency increased in deep layers, suggesting that routing behavior in deep layers is more severely affected by expert pruning. In contrast, under our proposed method, the selection frequency of each expert remained almost unchanged, as illustrated in the bottom-right figure in Figure \ref{fig:expert_usage}. This indicated that our approach largely preserved the original routing behavior. Maintaining stable expert routing is one of the key reasons for the effectiveness of our method.

\subsection{Error Analysis}
\label{sec:error_analysis}

Pruning the FFN layers inevitably caused discrepancies between the outputs of the pruned model and those of the full-parameter model. In general, pruning more parameters led to larger output deviations. Since the output of one layer serves as the input to the next, these deviations further influence subsequent routing decisions, causing errors to accumulate progressively across layers, as shown in Figure \ref{fig:diff_comparison}.
To quantify this effect, we computed the difference between the output vectors of the pruned model and the full model at each layer, and used the $\ell_2$ norm (i.e., Euclidean distance) of the difference vector as a measure of layer-wise error. Our method exhibited a similar overall error trend to the expert pruning method EASY-EP.
When ENP is applied in isolation, relatively large errors can be observed even in the shallow layers. This behavior arose because certain experts play a disproportionately important role; uniform neuron pruning removes neurons indiscriminately from both critical and less critical experts, which introduced a substantial error at early stages. Nevertheless, because ENP preserved the original routing structure, error accumulation across layers proceeds at a slower rate than with direct expert pruning. Consequently, in the middle and deep layers of the model, the error introduced by expert pruning exceeded that caused by neuron pruning.
Interestingly, although the error gradually accumulates across layers, it drops sharply at the final layer. This observation highlights the strong representational capacity of high-level experts. Motivated by this phenomenon, our model adopts a trapezoidal parameter distribution, allocating more parameters to higher layers.

\subsection{Generalization Ability}
\label{sec:generalization}
In this section, we designed a set of generalization experiments and compared our method with other approaches. We performed pruning using data from a single domain,e.g., mathematics, code, or science, and then evaluated the pruned models on both in domain and out of domain benchmarks. The experimental results are summarised in Table \ref{tab:generalizationResult} and Table \ref{tab:AblationMMLU}. For in-domain pruning, our method consistently outperforms expert pruning approaches. More importantly, our method achieves the highest average performance among the compared methods on out of domain evaluation. We attribute this strong generalization capability to the fact that our approach better preserves the original routing behavior and the overall structural integrity of the model.

\begin{table*}[h!]
\centering
\resizebox{0.75\textwidth}{!}{
\begin{tabular}{ll*{5}{c}}
\toprule
Method &Param &Mem(GB) &Input &Output &Total &Rate \\
\midrule
Full & 100\% & 32.95 & 2790.43 & 2793.08 & 5583.51 & 1.0   \\ 
Expert Pruning & 50\% & 18.55 & 3941.07 & 3944.81 & 7885.88 & 1.41   \\ 
ENP(Ours) & 50\% & \textbf{16.37} & \textbf{4094.03} & \textbf{4097.91} & \textbf{8191.94} & \textbf{1.47}   \\ 
\bottomrule
\end{tabular}
}
\caption{A comparison of throughput and static memory consumption among the full-parameter model, expert pruning methods, and our method.
Param denotes the routed expert retention ratio, and Mem represents the static GPU memory consumption of the Qwen model on a single A100-SXM-80GB GPU. Input denotes the input token throughput (tok/s), Output denotes the output token throughput (tok/s), Total denotes the total token throughput (tok/s), and Rate represents the ratio of the total throughput of each method relative to that of the full-parameter model.
}
\label{tab:MemThroughput}
\end{table*}

\subsection{Static Memory Consumption and Throughput}

We further compare our ENP method with expert pruning approaches in terms of memory consumption and throughput.
As shown in Table~\ref{tab:MemThroughput}, under the same parameter scale (50\% routed expert sparsity), ENP yields smaller individual experts, making it more friendly to GPU memory allocation, and significantly reduces static memory consumption from 32.95\,GB to 16.37\,GB.
With fewer activated parameters, ENP achieves higher input, output, and total throughput, increasing the total throughput to 147\% relative to the full-parameter model, and outperforming the expert pruning model with the same parameter scale by 6\%.

Unlike unstructured pruning methods, both ENP and TENP do not impose any special hardware requirements.
TENP requires modifications to the inference framework (e.g., vLLM, SGLang) to accommodate experts of different sizes, whereas ENP can be deployed without modifying the inference framework.

\begin{figure}[h!]
\centering
  \includegraphics[width=\columnwidth]{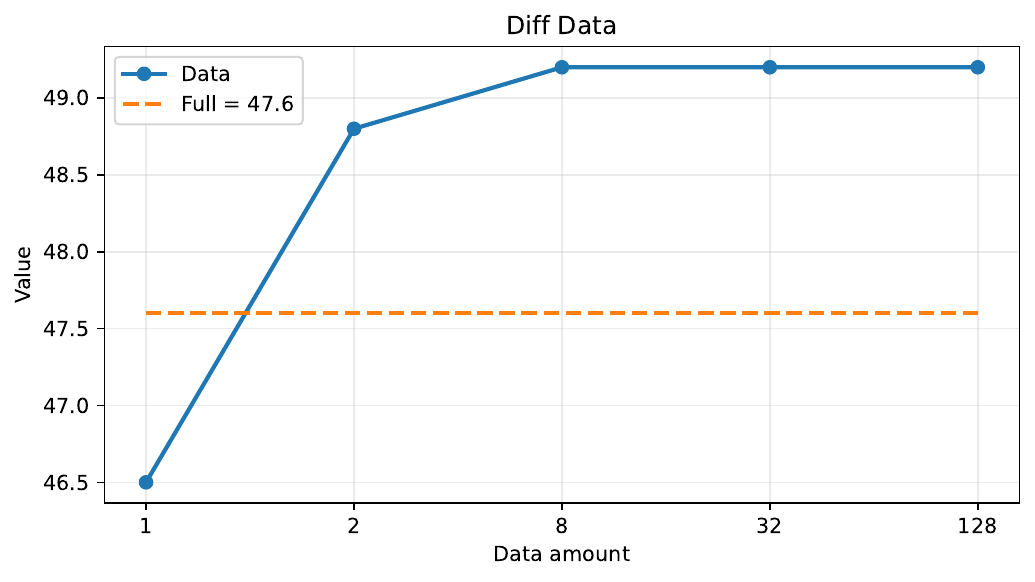}
  \caption{The impact of different data scales on pruning performance.}
  \label{fig:diff_data}
\end{figure}

\subsection{The Impact of Different Data Scales on Pruning Performance}
\label{sec:data}

Following the data selection strategy in EASY-EP and considering practical scenarios where sufficient samples may not be available, we use only the prompts from the dataset and the responses generated by the full-parameter model itself, thereby minimizing human involvement.
We further investigate the impact of different data scales on model performance, as shown in Figure~\ref{fig:diff_data}.
For a single application scenario, under a routed expert sparsity of 60\%, using only one sample is sufficient for the model to retain most of its performance.
With two samples, the pruned model can already surpass the performance of the full-parameter model.
Performance plateaus at eight samples, and further scaling brings no significant improvement.

\section{Conclusion}

In this paper, we have presented TENP, a pruning method for MoE models that preserves important experts while structurally pruning unimportant neurons within experts. Extensive experiments demonstrate that our approach better maintains the original routing behavior of the model, induces smaller intermediate-layer errors, and achieves superior generalization performance. Moreover, TENP consistently performs well across different sparsity levels, model architectures, and benchmarks.

\section*{Limitations}

Although TENP has been shown to be highly effective on Qwen and DeepSeek models, and larger models are expected to contain more redundant parameters suggesting that TENP could yield even greater pruning benefits we have not yet conducted experiments on extremely large scale mixture-of-experts models such as DeepSeek-V3.2. We leave the evaluation of TENP on such large-scale models to future work.

\section*{Acknowledgments}

The present research was supported by the National Key Research and Development Program of China (Grant No. 2023YFE0116400). We would like to thank the anonymous reviewers for their insightful comments.


\bibliography{custom}

@misc{DiEP,
      title={DiEP: Adaptive Mixture-of-Experts Compression through Differentiable Expert Pruning}, 
      author={Sikai Bai and Haoxi Li and Jie Zhang and Zicong Hong and Song Guo},
      year={2025},
      eprint={2509.16105},
      archivePrefix={arXiv},
      primaryClass={cs.CL},
      url={https://arxiv.org/abs/2509.16105}, 
}

@misc{GSM8K,
      title={Training Verifiers to Solve Math Word Problems}, 
      author={Karl Cobbe and Vineet Kosaraju and Mohammad Bavarian and Mark Chen and Heewoo Jun and Lukasz Kaiser and Matthias Plappert and Jerry Tworek and Jacob Hilton and Reiichiro Nakano and Christopher Hesse and John Schulman},
      year={2021},
      eprint={2110.14168},
      archivePrefix={arXiv},
      primaryClass={cs.LG},
      url={https://arxiv.org/abs/2110.14168}, 
}

@misc{MBPP,
      title={Program Synthesis with Large Language Models}, 
      author={Jacob Austin and Augustus Odena and Maxwell Nye and Maarten Bosma and Henryk Michalewski and David Dohan and Ellen Jiang and Carrie Cai and Michael Terry and Quoc Le and Charles Sutton},
      year={2021},
      eprint={2108.07732},
      archivePrefix={arXiv},
      primaryClass={cs.PL},
      url={https://arxiv.org/abs/2108.07732}, 
}

@misc{HumanEval,
      title={Evaluating Large Language Models Trained on Code}, 
      author={Mark Chen and Jerry Tworek and Heewoo Jun and Qiming Yuan and Henrique Ponde de Oliveira Pinto and Jared Kaplan and Harri Edwards and Yuri Burda and Nicholas Joseph and Greg Brockman and Alex Ray and Raul Puri and Gretchen Krueger and Michael Petrov and Heidy Khlaaf and Girish Sastry and Pamela Mishkin and Brooke Chan and Scott Gray and Nick Ryder and Mikhail Pavlov and Alethea Power and Lukasz Kaiser and Mohammad Bavarian and Clemens Winter and Philippe Tillet and Felipe Petroski Such and Dave Cummings and Matthias Plappert and Fotios Chantzis and Elizabeth Barnes and Ariel Herbert-Voss and William Hebgen Guss and Alex Nichol and Alex Paino and Nikolas Tezak and Jie Tang and Igor Babuschkin and Suchir Balaji and Shantanu Jain and William Saunders and Christopher Hesse and Andrew N. Carr and Jan Leike and Josh Achiam and Vedant Misra and Evan Morikawa and Alec Radford and Matthew Knight and Miles Brundage and Mira Murati and Katie Mayer and Peter Welinder and Bob McGrew and Dario Amodei and Sam McCandlish and Ilya Sutskever and Wojciech Zaremba},
      year={2021},
      eprint={2107.03374},
      archivePrefix={arXiv},
      primaryClass={cs.LG},
      url={https://arxiv.org/abs/2107.03374}, 
}

@misc{ARC,
      title={Think you have Solved Question Answering? Try ARC, the AI2 Reasoning Challenge}, 
      author={Peter Clark and Isaac Cowhey and Oren Etzioni and Tushar Khot and Ashish Sabharwal and Carissa Schoenick and Oyvind Tafjord},
      year={2018},
      eprint={1803.05457},
      archivePrefix={arXiv},
      primaryClass={cs.AI},
      url={https://arxiv.org/abs/1803.05457}, 
}

@inproceedings{EAC-MoE,
    title = "{EAC}-{M}o{E}: Expert-Selection Aware Compressor for Mixture-of-Experts Large Language Models",
    author = "Chen, Yuanteng  and
      Shao, Yuantian  and
      Wang, Peisong  and
      Cheng, Jian",
    editor = "Che, Wanxiang  and
      Nabende, Joyce  and
      Shutova, Ekaterina  and
      Pilehvar, Mohammad Taher",
    booktitle = "Proceedings of the 63rd Annual Meeting of the Association for Computational Linguistics (Volume 1: Long Papers)",
    month = jul,
    year = "2025",
    address = "Vienna, Austria",
    publisher = "Association for Computational Linguistics",
    url = "https://aclanthology.org/2025.acl-long.633/",
    doi = "10.18653/v1/2025.acl-long.633",
    pages = "12942--12963",
    ISBN = "979-8-89176-251-0"
}

@misc{Qwen3,
      title={Qwen3 Technical Report}, 
      author={An Yang and Anfeng Li and Baosong Yang and Beichen Zhang and Binyuan Hui and Bo Zheng and Bowen Yu and Chang Gao and Chengen Huang and Chenxu Lv and Chujie Zheng and Dayiheng Liu and Fan Zhou and Fei Huang and Feng Hu and Hao Ge and Haoran Wei and Huan Lin and Jialong Tang and Jian Yang and Jianhong Tu and Jianwei Zhang and Jianxin Yang and Jiaxi Yang and Jing Zhou and Jingren Zhou and Junyang Lin and Kai Dang and Keqin Bao and Kexin Yang and Le Yu and Lianghao Deng and Mei Li and Mingfeng Xue and Mingze Li and Pei Zhang and Peng Wang and Qin Zhu and Rui Men and Ruize Gao and Shixuan Liu and Shuang Luo and Tianhao Li and Tianyi Tang and Wenbiao Yin and Xingzhang Ren and Xinyu Wang and Xinyu Zhang and Xuancheng Ren and Yang Fan and Yang Su and Yichang Zhang and Yinger Zhang and Yu Wan and Yuqiong Liu and Zekun Wang and Zeyu Cui and Zhenru Zhang and Zhipeng Zhou and Zihan Qiu},
      year={2025},
      eprint={2505.09388},
      archivePrefix={arXiv},
      primaryClass={cs.CL},
      url={https://arxiv.org/abs/2505.09388}, 
}

@misc{FirstMoE,
      title={Outrageously Large Neural Networks: The Sparsely-Gated Mixture-of-Experts Layer}, 
      author={Noam Shazeer and Azalia Mirhoseini and Krzysztof Maziarz and Andy Davis and Quoc Le and Geoffrey Hinton and Jeff Dean},
      year={2017},
      eprint={1701.06538},
      archivePrefix={arXiv},
      primaryClass={cs.LG},
      url={https://arxiv.org/abs/1701.06538}, 
}

@misc{SwitchTransformer,
      title={Switch Transformers: Scaling to Trillion Parameter Models with Simple and Efficient Sparsity}, 
      author={William Fedus and Barret Zoph and Noam Shazeer},
      year={2022},
      eprint={2101.03961},
      archivePrefix={arXiv},
      primaryClass={cs.LG},
      url={https://arxiv.org/abs/2101.03961}, 
}

@misc{GShard,
      title={GShard: Scaling Giant Models with Conditional Computation and Automatic Sharding}, 
      author={Dmitry Lepikhin and HyoukJoong Lee and Yuanzhong Xu and Dehao Chen and Orhan Firat and Yanping Huang and Maxim Krikun and Noam Shazeer and Zhifeng Chen},
      year={2020},
      eprint={2006.16668},
      archivePrefix={arXiv},
      primaryClass={cs.CL},
      url={https://arxiv.org/abs/2006.16668}, 
}

@misc{DeepseekMoE,
      title={DeepSeekMoE: Towards Ultimate Expert Specialization in Mixture-of-Experts Language Models}, 
      author={Damai Dai and Chengqi Deng and Chenggang Zhao and R. X. Xu and Huazuo Gao and Deli Chen and Jiashi Li and Wangding Zeng and Xingkai Yu and Y. Wu and Zhenda Xie and Y. K. Li and Panpan Huang and Fuli Luo and Chong Ruan and Zhifang Sui and Wenfeng Liang},
      year={2024},
      eprint={2401.06066},
      archivePrefix={arXiv},
      primaryClass={cs.CL},
      url={https://arxiv.org/abs/2401.06066}, 
}

@misc{Qwen1.5MoE-A2.7B,
    title = {Qwen1.5-MoE: Matching 7B Model Performance with 1/3 Activated Parameters"},
    url = {https://qwenlm.github.io/blog/qwen-moe/},
    author = {Qwen Team},
    month = {February},
    year = {2024}
}

@misc{DeepSeekV2Lite,
      title={DeepSeek-V2: A Strong, Economical, and Efficient Mixture-of-Experts Language Model}, 
      author={DeepSeek-AI and Aixin Liu and Bei Feng and Bin Wang and Bingxuan Wang and Bo Liu and Chenggang Zhao and Chengqi Dengr and Chong Ruan and Damai Dai and Daya Guo and Dejian Yang and Deli Chen and Dongjie Ji and Erhang Li and Fangyun Lin and Fuli Luo and Guangbo Hao and Guanting Chen and Guowei Li and H. Zhang and Hanwei Xu and Hao Yang and Haowei Zhang and Honghui Ding and Huajian Xin and Huazuo Gao and Hui Li and Hui Qu and J. L. Cai and Jian Liang and Jianzhong Guo and Jiaqi Ni and Jiashi Li and Jin Chen and Jingyang Yuan and Junjie Qiu and Junxiao Song and Kai Dong and Kaige Gao and Kang Guan and Lean Wang and Lecong Zhang and Lei Xu and Leyi Xia and Liang Zhao and Liyue Zhang and Meng Li and Miaojun Wang and Mingchuan Zhang and Minghua Zhang and Minghui Tang and Mingming Li and Ning Tian and Panpan Huang and Peiyi Wang and Peng Zhang and Qihao Zhu and Qinyu Chen and Qiushi Du and R. J. Chen and R. L. Jin and Ruiqi Ge and Ruizhe Pan and Runxin Xu and Ruyi Chen and S. S. Li and Shanghao Lu and Shangyan Zhou and Shanhuang Chen and Shaoqing Wu and Shengfeng Ye and Shirong Ma and Shiyu Wang and Shuang Zhou and Shuiping Yu and Shunfeng Zhou and Size Zheng and T. Wang and Tian Pei and Tian Yuan and Tianyu Sun and W. L. Xiao and Wangding Zeng and Wei An and Wen Liu and Wenfeng Liang and Wenjun Gao and Wentao Zhang and X. Q. Li and Xiangyue Jin and Xianzu Wang and Xiao Bi and Xiaodong Liu and Xiaohan Wang and Xiaojin Shen and Xiaokang Chen and Xiaosha Chen and Xiaotao Nie and Xiaowen Sun and Xiaoxiang Wang and Xin Liu and Xin Xie and Xingkai Yu and Xinnan Song and Xinyi Zhou and Xinyu Yang and Xuan Lu and Xuecheng Su and Y. Wu and Y. K. Li and Y. X. Wei and Y. X. Zhu and Yanhong Xu and Yanping Huang and Yao Li and Yao Zhao and Yaofeng Sun and Yaohui Li and Yaohui Wang and Yi Zheng and Yichao Zhang and Yiliang Xiong and Yilong Zhao and Ying He and Ying Tang and Yishi Piao and Yixin Dong and Yixuan Tan and Yiyuan Liu and Yongji Wang and Yongqiang Guo and Yuchen Zhu and Yuduan Wang and Yuheng Zou and Yukun Zha and Yunxian Ma and Yuting Yan and Yuxiang You and Yuxuan Liu and Z. Z. Ren and Zehui Ren and Zhangli Sha and Zhe Fu and Zhen Huang and Zhen Zhang and Zhenda Xie and Zhewen Hao and Zhihong Shao and Zhiniu Wen and Zhipeng Xu and Zhongyu Zhang and Zhuoshu Li and Zihan Wang and Zihui Gu and Zilin Li and Ziwei Xie},
      year={2024},
      eprint={2405.04434},
      archivePrefix={arXiv},
      primaryClass={cs.CL},
      url={https://arxiv.org/abs/2405.04434}, 
}

@misc{Mixtral8x7B,
      title={Mixtral of Experts}, 
      author={Albert Q. Jiang and Alexandre Sablayrolles and Antoine Roux and Arthur Mensch and Blanche Savary and Chris Bamford and Devendra Singh Chaplot and Diego de las Casas and Emma Bou Hanna and Florian Bressand and Gianna Lengyel and Guillaume Bour and Guillaume Lample and Lélio Renard Lavaud and Lucile Saulnier and Marie-Anne Lachaux and Pierre Stock and Sandeep Subramanian and Sophia Yang and Szymon Antoniak and Teven Le Scao and Théophile Gervet and Thibaut Lavril and Thomas Wang and Timothée Lacroix and William El Sayed},
      year={2024},
      eprint={2401.04088},
      archivePrefix={arXiv},
      primaryClass={cs.LG},
      url={https://arxiv.org/abs/2401.04088}, 
}

@misc{GPT-OSS,
      title={gpt-oss-120b \& gpt-oss-20b Model Card},
      author={OpenAI},
      year={2025},
      eprint={2508.10925},
      archivePrefix={arXiv},
      primaryClass={cs.CL},
      url={https://arxiv.org/abs/2508.10925},
}

@misc{DeepseekV3.2,
      title={DeepSeek-V3.2: Pushing the Frontier of Open Large Language Models}, 
      author={DeepSeek-AI and Aixin Liu and Aoxue Mei and Bangcai Lin and Bing Xue and Bingxuan Wang and Bingzheng Xu and Bochao Wu and Bowei Zhang and Chaofan Lin and Chen Dong and Chengda Lu and Chenggang Zhao and Chengqi Deng and Chenhao Xu and Chong Ruan and Damai Dai and Daya Guo and Dejian Yang and Deli Chen and Erhang Li and Fangqi Zhou and Fangyun Lin and Fucong Dai and Guangbo Hao and Guanting Chen and Guowei Li and H. Zhang and Hanwei Xu and Hao Li and Haofen Liang and Haoran Wei and Haowei Zhang and Haowen Luo and Haozhe Ji and Honghui Ding and Hongxuan Tang and Huanqi Cao and Huazuo Gao and Hui Qu and Hui Zeng and Jialiang Huang and Jiashi Li and Jiaxin Xu and Jiewen Hu and Jingchang Chen and Jingting Xiang and Jingyang Yuan and Jingyuan Cheng and Jinhua Zhu and Jun Ran and Junguang Jiang and Junjie Qiu and Junlong Li and Junxiao Song and Kai Dong and Kaige Gao and Kang Guan and Kexin Huang and Kexing Zhou and Kezhao Huang and Kuai Yu and Lean Wang and Lecong Zhang and Lei Wang and Liang Zhao and Liangsheng Yin and Lihua Guo and Lingxiao Luo and Linwang Ma and Litong Wang and Liyue Zhang and M. S. Di and M. Y Xu and Mingchuan Zhang and Minghua Zhang and Minghui Tang and Mingxu Zhou and Panpan Huang and Peixin Cong and Peiyi Wang and Qiancheng Wang and Qihao Zhu and Qingyang Li and Qinyu Chen and Qiushi Du and Ruiling Xu and Ruiqi Ge and Ruisong Zhang and Ruizhe Pan and Runji Wang and Runqiu Yin and Runxin Xu and Ruomeng Shen and Ruoyu Zhang and S. H. Liu and Shanghao Lu and Shangyan Zhou and Shanhuang Chen and Shaofei Cai and Shaoyuan Chen and Shengding Hu and Shengyu Liu and Shiqiang Hu and Shirong Ma and Shiyu Wang and Shuiping Yu and Shunfeng Zhou and Shuting Pan and Songyang Zhou and Tao Ni and Tao Yun and Tian Pei and Tian Ye and Tianyuan Yue and Wangding Zeng and Wen Liu and Wenfeng Liang and Wenjie Pang and Wenjing Luo and Wenjun Gao and Wentao Zhang and Xi Gao and Xiangwen Wang and Xiao Bi and Xiaodong Liu and Xiaohan Wang and Xiaokang Chen and Xiaokang Zhang and Xiaotao Nie and Xin Cheng and Xin Liu and Xin Xie and Xingchao Liu and Xingkai Yu and Xingyou Li and Xinyu Yang and Xinyuan Li and Xu Chen and Xuecheng Su and Xuehai Pan and Xuheng Lin and Xuwei Fu and Y. Q. Wang and Yang Zhang and Yanhong Xu and Yanru Ma and Yao Li and Yao Li and Yao Zhao and Yaofeng Sun and Yaohui Wang and Yi Qian and Yi Yu and Yichao Zhang and Yifan Ding and Yifan Shi and Yiliang Xiong and Ying He and Ying Zhou and Yinmin Zhong and Yishi Piao and Yisong Wang and Yixiao Chen and Yixuan Tan and Yixuan Wei and Yiyang Ma and Yiyuan Liu and Yonglun Yang and Yongqiang Guo and Yongtong Wu and Yu Wu and Yuan Cheng and Yuan Ou and Yuanfan Xu and Yuduan Wang and Yue Gong and Yuhan Wu and Yuheng Zou and Yukun Li and Yunfan Xiong and Yuxiang Luo and Yuxiang You and Yuxuan Liu and Yuyang Zhou and Z. F. Wu and Z. Z. Ren and Zehua Zhao and Zehui Ren and Zhangli Sha and Zhe Fu and Zhean Xu and Zhenda Xie and Zhengyan Zhang and Zhewen Hao and Zhibin Gou and Zhicheng Ma and Zhigang Yan and Zhihong Shao and Zhixian Huang and Zhiyu Wu and Zhuoshu Li and Zhuping Zhang and Zian Xu and Zihao Wang and Zihui Gu and Zijia Zhu and Zilin Li and Zipeng Zhang and Ziwei Xie and Ziyi Gao and Zizheng Pan and Zongqing Yao and Bei Feng and Hui Li and J. L. Cai and Jiaqi Ni and Lei Xu and Meng Li and Ning Tian and R. J. Chen and R. L. Jin and S. S. Li and Shuang Zhou and Tianyu Sun and X. Q. Li and Xiangyue Jin and Xiaojin Shen and Xiaosha Chen and Xinnan Song and Xinyi Zhou and Y. X. Zhu and Yanping Huang and Yaohui Li and Yi Zheng and Yuchen Zhu and Yunxian Ma and Zhen Huang and Zhipeng Xu and Zhongyu Zhang and Dongjie Ji and Jian Liang and Jianzhong Guo and Jin Chen and Leyi Xia and Miaojun Wang and Mingming Li and Peng Zhang and Ruyi Chen and Shangmian Sun and Shaoqing Wu and Shengfeng Ye and T. Wang and W. L. Xiao and Wei An and Xianzu Wang and Xiaowen Sun and Xiaoxiang Wang and Ying Tang and Yukun Zha and Zekai Zhang and Zhe Ju and Zhen Zhang and Zihua Qu},
      year={2025},
      eprint={2512.02556},
      archivePrefix={arXiv},
      primaryClass={cs.CL},
      url={https://arxiv.org/abs/2512.02556}, 
}

@misc{KimiK2,
      title={Kimi K2: Open Agentic Intelligence}, 
      author={Kimi Team and Yifan Bai and Yiping Bao and Guanduo Chen and Jiahao Chen and Ningxin Chen and Ruijue Chen and Yanru Chen and Yuankun Chen and Yutian Chen and Zhuofu Chen and Jialei Cui and Hao Ding and Mengnan Dong and Angang Du and Chenzhuang Du and Dikang Du and Yulun Du and Yu Fan and Yichen Feng and Kelin Fu and Bofei Gao and Hongcheng Gao and Peizhong Gao and Tong Gao and Xinran Gu and Longyu Guan and Haiqing Guo and Jianhang Guo and Hao Hu and Xiaoru Hao and Tianhong He and Weiran He and Wenyang He and Chao Hong and Yangyang Hu and Zhenxing Hu and Weixiao Huang and Zhiqi Huang and Zihao Huang and Tao Jiang and Zhejun Jiang and Xinyi Jin and Yongsheng Kang and Guokun Lai and Cheng Li and Fang Li and Haoyang Li and Ming Li and Wentao Li and Yanhao Li and Yiwei Li and Zhaowei Li and Zheming Li and Hongzhan Lin and Xiaohan Lin and Zongyu Lin and Chengyin Liu and Chenyu Liu and Hongzhang Liu and Jingyuan Liu and Junqi Liu and Liang Liu and Shaowei Liu and T. Y. Liu and Tianwei Liu and Weizhou Liu and Yangyang Liu and Yibo Liu and Yiping Liu and Yue Liu and Zhengying Liu and Enzhe Lu and Lijun Lu and Shengling Ma and Xinyu Ma and Yingwei Ma and Shaoguang Mao and Jie Mei and Xin Men and Yibo Miao and Siyuan Pan and Yebo Peng and Ruoyu Qin and Bowen Qu and Zeyu Shang and Lidong Shi and Shengyuan Shi and Feifan Song and Jianlin Su and Zhengyuan Su and Xinjie Sun and Flood Sung and Heyi Tang and Jiawen Tao and Qifeng Teng and Chensi Wang and Dinglu Wang and Feng Wang and Haiming Wang and Jianzhou Wang and Jiaxing Wang and Jinhong Wang and Shengjie Wang and Shuyi Wang and Yao Wang and Yejie Wang and Yiqin Wang and Yuxin Wang and Yuzhi Wang and Zhaoji Wang and Zhengtao Wang and Zhexu Wang and Chu Wei and Qianqian Wei and Wenhao Wu and Xingzhe Wu and Yuxin Wu and Chenjun Xiao and Xiaotong Xie and Weimin Xiong and Boyu Xu and Jing Xu and Jinjing Xu and L. H. Xu and Lin Xu and Suting Xu and Weixin Xu and Xinran Xu and Yangchuan Xu and Ziyao Xu and Junjie Yan and Yuzi Yan and Xiaofei Yang and Ying Yang and Zhen Yang and Zhilin Yang and Zonghan Yang and Haotian Yao and Xingcheng Yao and Wenjie Ye and Zhuorui Ye and Bohong Yin and Longhui Yu and Enming Yuan and Hongbang Yuan and Mengjie Yuan and Haobing Zhan and Dehao Zhang and Hao Zhang and Wanlu Zhang and Xiaobin Zhang and Yangkun Zhang and Yizhi Zhang and Yongting Zhang and Yu Zhang and Yutao Zhang and Yutong Zhang and Zheng Zhang and Haotian Zhao and Yikai Zhao and Huabin Zheng and Shaojie Zheng and Jianren Zhou and Xinyu Zhou and Zaida Zhou and Zhen Zhu and Weiyu Zhuang and Xinxing Zu},
      year={2025},
      eprint={2507.20534},
      archivePrefix={arXiv},
      primaryClass={cs.LG},
      url={https://arxiv.org/abs/2507.20534}, 
}

@misc{DA-MoE,
      title={DA-MoE: Addressing Depth-Sensitivity in Graph-Level Analysis through Mixture of Experts}, 
      author={Zelin Yao and Chuang Liu and Xianke Meng and Yibing Zhan and Jia Wu and Shirui Pan and Wenbin Hu},
      year={2024},
      eprint={2411.03025},
      archivePrefix={arXiv},
      primaryClass={cs.LG},
      url={https://arxiv.org/abs/2411.03025}, 
}

@misc{XMoE,
      title={XMoE: Sparse Models with Fine-grained and Adaptive Expert Selection}, 
      author={Yuanhang Yang and Shiyi Qi and Wenchao Gu and Chaozheng Wang and Cuiyun Gao and Zenglin Xu},
      year={2024},
      eprint={2403.18926},
      archivePrefix={arXiv},
      primaryClass={cs.LG},
      url={https://arxiv.org/abs/2403.18926}, 
}

@misc{MOLA,
      title={Higher Layers Need More LoRA Experts}, 
      author={Chongyang Gao and Kezhen Chen and Jinmeng Rao and Baochen Sun and Ruibo Liu and Daiyi Peng and Yawen Zhang and Xiaoyuan Guo and Jie Yang and VS Subrahmanian},
      year={2024},
      eprint={2402.08562},
      archivePrefix={arXiv},
      primaryClass={cs.CL},
      url={https://arxiv.org/abs/2402.08562}, 
}

@misc{GroveMoE,
      title={Grove MoE: Towards Efficient and Superior MoE LLMs with Adjugate Experts}, 
      author={Haoyuan Wu and Haoxing Chen and Xiaodong Chen and Zhanchao Zhou and Tieyuan Chen and Yihong Zhuang and Guoshan Lu and Zenan Huang and Junbo Zhao and Lin Liu and Zhenzhong Lan and Bei Yu and Jianguo Li},
      year={2025},
      eprint={2508.07785},
      archivePrefix={arXiv},
      primaryClass={cs.CL},
      url={https://arxiv.org/abs/2508.07785}, 
}

@misc{SEER-MoE,
      title={SEER-MoE: Sparse Expert Efficiency through Regularization for Mixture-of-Experts}, 
      author={Alexandre Muzio and Alex Sun and Churan He},
      year={2024},
      eprint={2404.05089},
      archivePrefix={arXiv},
      primaryClass={cs.CL},
      url={https://arxiv.org/abs/2404.05089}, 
}

@misc{Wanda,
      title={A Simple and Effective Pruning Approach for Large Language Models}, 
      author={Mingjie Sun and Zhuang Liu and Anna Bair and J. Zico Kolter},
      year={2024},
      eprint={2306.11695},
      archivePrefix={arXiv},
      primaryClass={cs.CL},
      url={https://arxiv.org/abs/2306.11695}, 
}

@misc{NotAllExpertsAreEqual,
      title={Not All Experts are Equal: Efficient Expert Pruning and Skipping for Mixture-of-Experts Large Language Models}, 
      author={Xudong Lu and Qi Liu and Yuhui Xu and Aojun Zhou and Siyuan Huang and Bo Zhang and Junchi Yan and Hongsheng Li},
      year={2024},
      eprint={2402.14800},
      archivePrefix={arXiv},
      primaryClass={cs.CL},
      url={https://arxiv.org/abs/2402.14800}, 
}

@misc{DomainPruning,
      title={Domain-Specific Pruning of Large Mixture-of-Experts Models with Few-shot Demonstrations}, 
      author={Zican Dong and Han Peng and Peiyu Liu and Wayne Xin Zhao and Dong Wu and Feng Xiao and Zhifeng Wang},
      year={2025},
      eprint={2504.06792},
      archivePrefix={arXiv},
      primaryClass={cs.CL},
      url={https://arxiv.org/abs/2504.06792}, 
}

@misc{NeuronExperts,
      title={Mixture of Neuron Experts}, 
      author={Runxi Cheng and Yuchen Guan and Yucheng Ding and Qingguo Hu and Yongxian Wei and Chun Yuan and Yelong Shen and Weizhu Chen and Yeyun Gong},
      year={2025},
      eprint={2510.05781},
      archivePrefix={arXiv},
      primaryClass={cs.CL},
      url={https://arxiv.org/abs/2510.05781}, 
}

@misc{MoE-I2,
      title={MoE-I$^2$: Compressing Mixture of Experts Models through Inter-Expert Pruning and Intra-Expert Low-Rank Decomposition}, 
      author={Cheng Yang and Yang Sui and Jinqi Xiao and Lingyi Huang and Yu Gong and Yuanlin Duan and Wenqi Jia and Miao Yin and Yu Cheng and Bo Yuan},
      year={2024},
      eprint={2411.01016},
      archivePrefix={arXiv},
      primaryClass={cs.LG},
      url={https://arxiv.org/abs/2411.01016}, 
}

@misc{TaskPurning,
      title={Task-Specific Expert Pruning for Sparse Mixture-of-Experts}, 
      author={Tianyu Chen and Shaohan Huang and Yuan Xie and Binxing Jiao and Daxin Jiang and Haoyi Zhou and Jianxin Li and Furu Wei},
      year={2022},
      eprint={2206.00277},
      archivePrefix={arXiv},
      primaryClass={cs.LG},
      url={https://arxiv.org/abs/2206.00277}, 
}

@misc{SparseGPT,
      title={SparseGPT: Massive Language Models Can Be Accurately Pruned in One-Shot}, 
      author={Elias Frantar and Dan Alistarh},
      year={2023},
      eprint={2301.00774},
      archivePrefix={arXiv},
      primaryClass={cs.LG},
      url={https://arxiv.org/abs/2301.00774}, 
}

@misc{MoE-Pruner,
      title={MoE-Pruner: Pruning Mixture-of-Experts Large Language Model using the Hints from Its Router}, 
      author={Yanyue Xie and Zhi Zhang and Ding Zhou and Cong Xie and Ziang Song and Xin Liu and Yanzhi Wang and Xue Lin and An Xu},
      year={2024},
      eprint={2410.12013},
      archivePrefix={arXiv},
      primaryClass={cs.CL},
      url={https://arxiv.org/abs/2410.12013}, 
}

@misc{NeuronMerge,
      title={Efficient Expert Pruning for Sparse Mixture-of-Experts Language Models: Enhancing Performance and Reducing Inference Costs}, 
      author={Enshu Liu and Junyi Zhu and Zinan Lin and Xuefei Ning and Matthew B. Blaschko and Shengen Yan and Guohao Dai and Huazhong Yang and Yu Wang},
      year={2024},
      eprint={2407.00945},
      archivePrefix={arXiv},
      primaryClass={cs.LG},
      url={https://arxiv.org/abs/2407.00945}, 
}

@misc{MergeThenCompress,
      title={Merge, Then Compress: Demystify Efficient SMoE with Hints from Its Routing Policy}, 
      author={Pingzhi Li and Zhenyu Zhang and Prateek Yadav and Yi-Lin Sung and Yu Cheng and Mohit Bansal and Tianlong Chen},
      year={2024},
      eprint={2310.01334},
      archivePrefix={arXiv},
      primaryClass={cs.LG},
      url={https://arxiv.org/abs/2310.01334}, 
}

@misc{MergeExpert,
      title={CoMoE: Collaborative Optimization of Expert Aggregation and Offloading for MoE-based LLMs at Edge}, 
      author={Muqing Li and Ning Li and Xin Yuan and Wenchao Xu and Quan Chen and Song Guo and Haijun Zhang},
      year={2025},
      eprint={2508.09208},
      archivePrefix={arXiv},
      primaryClass={cs.NI},
      url={https://arxiv.org/abs/2508.09208}, 
}

@misc{Merge2One,
      title={Merging Experts into One: Improving Computational Efficiency of Mixture of Experts}, 
      author={Shwai He and Run-Ze Fan and Liang Ding and Li Shen and Tianyi Zhou and Dacheng Tao},
      year={2023},
      eprint={2310.09832},
      archivePrefix={arXiv},
      primaryClass={cs.CL},
      url={https://arxiv.org/abs/2310.09832}, 
}

@misc{2Dense,
      title={Condense, Don't Just Prune: Enhancing Efficiency and Performance in MoE Layer Pruning}, 
      author={Mingyu Cao and Gen Li and Jie Ji and Jiaqi Zhang and Xiaolong Ma and Shiwei Liu and Lu Yin},
      year={2025},
      eprint={2412.00069},
      archivePrefix={arXiv},
      primaryClass={cs.LG},
      url={https://arxiv.org/abs/2412.00069}, 
}

@inproceedings{QuantizationRenRen,
    title = "A Comprehensive Evaluation of Quantization Strategies for Large Language Models",
    author = "Jin, Renren  and
      Du, Jiangcun  and
      Huang, Wuwei  and
      Liu, Wei  and
      Luan, Jian  and
      Wang, Bin  and
      Xiong, Deyi",
    editor = "Ku, Lun-Wei  and
      Martins, Andre  and
      Srikumar, Vivek",
    booktitle = "Findings of the Association for Computational Linguistics: ACL 2024",
    month = aug,
    year = "2024",
    address = "Bangkok, Thailand",
    publisher = "Association for Computational Linguistics",
    url = "https://aclanthology.org/2024.findings-acl.726/",
    doi = "10.18653/v1/2024.findings-acl.726",
    pages = "12186--12215"
}

@inproceedings{QuantizationJiangcun,
    author = {Du, Jiangcun and Jin, Renren and Huang, Wuwei and Liu, Wei and Luan, Jian and Xiong, Deyi},
    title = {Optimize Quantization for Large Language Models via Progressive Training},
    year = {2025},
    isbn = {9798400713316},
    publisher = {Association for Computing Machinery},
    address = {New York, NY, USA},
    url = {https://doi.org/10.1145/3701716.3717578},
    doi = {10.1145/3701716.3717578},
    booktitle = {Companion Proceedings of the ACM on Web Conference 2025},
    pages = {2474–2483},
    numpages = {10},
    location = {Sydney NSW, Australia},
    series = {WWW '25}
}

@article{MultilingualLLM,
  title={Multilingual Large Language Models: A Systematic Survey},
  author={Shaolin Zhu and Supryadi and Shaoyang Xu and Haoran Sun and Leiyu Pan and Menglong Cui and Jiangcun Du and Renren Jin and Ant{\'o}nio Branco and Deyi Xiong},
  journal={ArXiv},
  year={2024},
  volume={abs/2411.11072},
  url={https://api.semanticscholar.org/CorpusID:274131470}
}

@inproceedings{fuxitranyu,
    title = "{F}uxi{T}ranyu: A Multilingual Large Language Model Trained with Balanced Data",
    author = "Sun, Haoran  and
      Jin, Renren  and
      Xu, Shaoyang  and
      Pan, Leiyu  and
      Supryadi  and
      Cui, Menglong  and
      Du, Jiangcun  and
      Lei, Yikun  and
      Yang, Lei  and
      Shi, Ling  and
      Xiao, Juesi  and
      Zhu, Shaolin  and
      Xiong, Deyi",
    editor = "Dernoncourt, Franck  and
      Preo{\c{t}}iuc-Pietro, Daniel  and
      Shimorina, Anastasia",
    booktitle = "Proceedings of the 2024 Conference on Empirical Methods in Natural Language Processing: Industry Track",
    month = nov,
    year = "2024",
    address = "Miami, Florida, US",
    publisher = "Association for Computational Linguistics",
    url = "https://aclanthology.org/2024.emnlp-industry.110/",
    doi = "10.18653/v1/2024.emnlp-industry.110",
    pages = "1499--1522"
}

@article{TibetanLLM,
  title={Advancing Large Language Models for Tibetan with Curated Data and Continual Pre-Training},
  author={Leiyu Pan and Bojian Xiong and Lei Yang and Renren Jin and Shaowei Zhang and Yue Chen and Ling Shi and Jiang Zhou and Junru Wu and Zhen D. Wang and Jianxiang Peng and Juesi Xiao and Tianyu Dong and Zhuowen Han and Zhuo Chen and Yuqi Ren and Deyi Xiong},
  journal={ArXiv},
  year={2025},
  volume={abs/2507.09205},
  url={https://api.semanticscholar.org/CorpusID:280137725}
}

@inproceedings{ShortGPT,
    title = "{S}hort{GPT}: Layers in Large Language Models are More Redundant Than You Expect",
    author = "Men, Xin  and
      Xu, Mingyu  and
      Zhang, Qingyu  and
      Yuan, Qianhao  and
      Wang, Bingning  and
      Lin, Hongyu  and
      Lu, Yaojie  and
      Han, Xianpei  and
      Chen, Weipeng",
    editor = "Che, Wanxiang  and
      Nabende, Joyce  and
      Shutova, Ekaterina  and
      Pilehvar, Mohammad Taher",
    booktitle = "Findings of the Association for Computational Linguistics: ACL 2025",
    month = jul,
    year = "2025",
    address = "Vienna, Austria",
    publisher = "Association for Computational Linguistics",
    url = "https://aclanthology.org/2025.findings-acl.1035/",
    doi = "10.18653/v1/2025.findings-acl.1035",
    pages = "20192--20204",
    ISBN = "979-8-89176-256-5"
}

\clearpage

\appendix

\begin{table*}[h!]
\centering
\resizebox{0.95\textwidth}{!}{
\begin{tabular}{ll*{6}{c}}
\toprule
Model &Retain &GSM8K &MBPP &Humaneval &ARC-E &ARC-C &Avg.\\
\midrule
\multirow{6}*{\makecell{Qwen1.5MoE-A2.7B}}&  0\% & 51.3  & 40.6  & 28.0  & 84.6  & 74.9  & 55.88   \\ 
~ & 10\% & 52.9  & 42.3  & 32.3  & 85.1  & 75.0  & 57.52   \\ 
~ & 20\% & 53.8  & 42.9  & \textbf{32.3}  & 85.1  & 74.6  & 57.74   \\ 
~ & 30\% & \textbf{58.3}  & \textbf{45.7}  & 31.1  & \textbf{85.1}  & \textbf{75.1}  & \textbf{59.06}   \\ 
~ & 40\% & 57.5  & 44.5  & 30.5  & 83.0  & 73.5  & 57.80   \\ 
~ & 50\% & 54.3  & 42.3  & 26.8  & 81.2  & 71.0  & 55.12   \\ 
\midrule
\multirow{6}*{\makecell{DeepSeek-V2-Lite}}& 0\% & 22.1  & 33.9  & 19.5  & 77.8  & 65.7  & 43.80   \\ 
~ & 10\% & 36.8  & \textbf{48.0}  & 28.7  & 78.1  & 64.7  & 51.26   \\ 
~ & 20\% & 38.4  & 45.7  & \textbf{29.9}  & \textbf{79.1}  & \textbf{66.8}  & \textbf{51.98}   \\ 
~ & 30\% & \textbf{38.7}  & 44.5  & 26.2  & 78.5  & 66.7  & 50.92   \\ 
~ & 40\% & 38.6  & 40.9  & 23.8  & 76.4  & 62.6  & 48.46   \\ 
~ & 50\% & 25.3  & 34.6  & 13.4  & 70.9  & 56.2  & 40.08   \\ 
\bottomrule
\end{tabular}
}
\caption{The performance of our TENP method on different datasets under varying important expert retention ratios, where Retain denotes the important expert retention ratio.}
\label{tab:RetainRatio}
\end{table*}

\section{Model Detail}
\label{sce:ModelDetail}
Qwen1.5-MoE-A2.7B contains 14 billion parameters and 24 layers.  
Each layer consists of 60 routed experts and 4 shared experts.  
For each token, the router selects the top 4 experts with the highest scores in each layer to perform the forward computation.DeepSeek-V2-Lite contains 16 billion parameters and 27 layers, where the first layer is a dense layer. Starting from the second layer, each layer includes 64 routed experts and 2 shared experts. For each token, the router selects the top 6 experts with the highest scores in each layer to perform the forward computation.

\section{Expert Neuron Importance Algorithm}
\label{sec:Algorithm}
Algorithm~\ref{alg:ENP} describes in detail our method for selecting important neurons within an expert.
Given only the parameters of a specific expert and the input vectors, we can compute the importance of each neuron based on this information.
By leveraging PyTorch’s broadcasting mechanism in matrix multiplication, we are able to compute the average importance of all neurons in an expert across all tokens using a single forward pass.

The algorithm illustrates that neuron importance is determined by computing the projection length of the output vector produced independently by each neuron onto the expert output, which is the superposition of the outputs of all neurons.
A simpler alternative is to directly evaluate the magnitude of each neuron’s output using its $L_2$ norm.
In this case, it suffices to directly compute the $L_2$ norm of $C$ in the algorithm.

\begin{algorithm}[h!]
\caption{Expert Neuron Importance Algorithm}
\label{alg:forward_simple}
\begin{algorithmic}
\Require $x \gets \textit{Input hidden states}$
\Comment $x: L\times d$
\Require $K \gets \textit{Neuron Number Of One Expert }$
\Require $W_G\gets \textit{Gate Matrix}$
\Comment $W_G:K\times d $
\Require $W_U\gets \textit{Up Matrix}$
\Comment $W_U:K\times d $
\Require $W_D\gets \textit{Down Matrix}$
\Comment $W_D:d\times K $
\Ensure $y \gets \textit{Output hidden states}$
\Comment $y: L\times d$
\Ensure $P \gets \textit{Neuron Importance}$
\Comment $P: K$

\State $m \gets \mathrm{act}(W_Gx \odot W_Ux)$
\Comment$m: L\times K$
\State $y \gets W_Dm$
\State $M \gets m^{\top}.\mathrm{unsqueeze}(-1)$
\Comment$M: K \times L \times 1$
\State $\mathbf{W}_{D2} \gets W_D^{\top}.\mathrm{unsqueeze}(1)$
\Comment$\mathbf{W}_{D2}: K \times 1 \times d$
\State $C \gets M \ @ \ \mathbf{W}_{D2}$
\Comment$C: K \times L \times d$

\State $Y \gets \textit{y}.\mathrm{unsqueeze}(0)$
\Comment$Y: 1 \times L \times d$
\State $s \gets (C \odot Y).\mathrm{sum}(\mathrm{dim}=-1)$
\Comment$s: K \times L$
\State $r \gets Y.\mathrm{norm}(p=2,\mathrm{dim}=-1)$
\Comment$r: K \times L$

\State $\varepsilon \gets 10^{-8}$
\State $P \gets s \ / \ (r + \varepsilon)$
\Comment$P: K \times L$
\State $P \gets P.\mathrm{mean}(\mathrm{dim}=1)$
\Comment$P: K$
\State \Return $y, P$
\end{algorithmic}
\label{alg:ENP}
\end{algorithm}

\begin{figure*}[h!]
\centering
  \includegraphics[width=\textwidth]{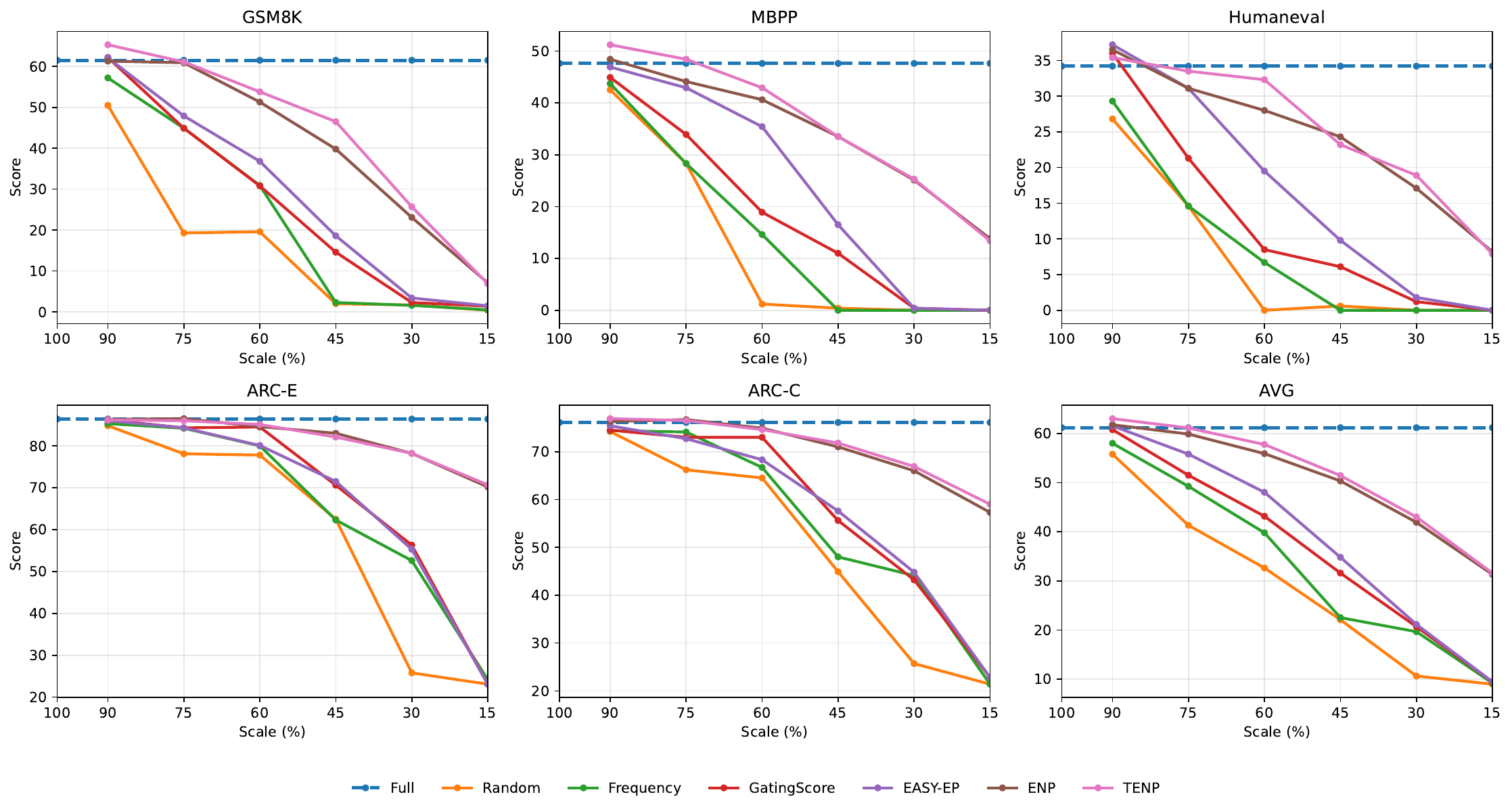}
  \caption{The performance of different model pruning methods across various benchmarks under different sparsity levels.}
  \label{fig:diff_scale}
\end{figure*}

\section{Pruning at different levels of sparsification}
\label{sec:DiffSparsity}
To more comprehensively reflect the effectiveness of our method under different sparsity levels of routed experts, we design experiments with scales of 15\%, 30\%, 45\%, 60\%, 75\%, and 90\%.
As shown in Figure~\ref{fig:diff_scale}, the experimental curves indicate that our method achieves the best performance on average across different sparsity levels.
At the scale of 90\%, our method outperforms the full-parameter model on nearly all benchmarks, with particularly significant improvements on code generation and mathematical reasoning tasks.
At the scale of 75\%, our method performs on par with the full-parameter model.
At the scale of 60\%, the model performance is slightly lower than that of the full-parameter model, yet still superior to other expert pruning methods.
At the scale of 45\%, 30\%, and 15\%, our method continues to preserve the core capabilities of the model.

\section{The Impact of Different Important Expert Retention Ratios on Pruning Performance}
\label{sec:ExpertRetentionRatios}
To evaluate the effect of different important expert retention ratios on pruning performance, we conduct experiments under a routed expert sparsity of 60\% with varying retention ratios to determine the optimal setting. As shown in Table~\ref{tab:RetainRatio}, the optimal important expert retention ratio for the Qwen model is 30\%, corresponding to a retention ratio of unimportant expert neurons of $(60\%-30\%)/(100\%-30\%) \approx 42.86\%$. For the DeepSeek model, the optimal important-expert retention ratio is 20\%, with the corresponding retention ratio of unimportant expert neurons being $(60\%-20\%)/(100\%-20\%) = 50.00\%$. When the important-expert retention ratio is 0\%, the TENP method degenerates into the ENP method. From the table, we observe that as the retention ratio increases, accuracy first improves and then degrades, with the optimal retention ratios concentrated in the middle range.
By default, setting the parameter budget of important experts equal to that of unimportant experts yields favorable performance.

\begin{table*}[h!]
\centering
\resizebox{0.95\textwidth}{!}{
    \begin{tabular}{llcccccccc}
    \toprule
        Model & Method & E & Total Sparsity & GSM8K & MBPP & HumanEval & ARC-E & ARC-C & Avg.  \\ 
    \midrule
        \multirow{5}*{\makecell{Qwen1.5MoE-A2.7B}}& ENP & 60\% & 35\% & 51.3 & 40.6 & 28.0  & 84.6 & 74.9 & 55.88   \\ 
        ~ & TENP+BI(MoDeGPT) & 60\% & 35\% & 49.1 & 39.4 & 30.5 & 85.2 & \textbf{75.3} & 55.90   \\ 
        ~ & TENP(Rectange) & 60\% & 35\% & 51.1 & 39.1 & 29.2 & 85.2 & \textbf{75.3} & 55.98   \\ 
        ~ & TENP+Rank(ShortGPT) & 60\% & 35\% & 49.7 & 41.3 & 28.0  & \textbf{85.7} & \textbf{75.3} & 56.00   \\ 
        ~ & TENP (Ours) & 60\% & 35\% & \textbf{58.3}  & \textbf{45.7}  & \textbf{31.1}  & 85.1  & 75.1  & \textbf{59.06}   \\ 
    \bottomrule
    \end{tabular}
}
\caption{A comparison of other layer selection methods with our method}
\label{tab:RetainRatio}
\end{table*}

\begin{table*}[h!]
\centering
\resizebox{0.95\textwidth}{!}{
    \begin{tabular}{llccccccc}
    \toprule
        Model & Method & E & GSM8K & MBPP & HumanEval & ARC-E & ARC-C & Avg.  \\ 
    \midrule
        \multirow{6}*{\makecell{Qwen3-Next-80B-A3B-Instruct}} & Full & 100.00\% & 93.7  & 76.7  & 84.1  & 94.8  & 93.9  & 88.64   \\ 
    \cmidrule(lr){2-9}
        ~ & Random & 60.00\% & 85.8  & 63.3  & 64.6  & 89.9  & 88.1  & 78.34   \\ 
        ~ & Frequency & 60.00\% & 87.8  & 72.2  & 73.1  & 94.0  & 91.5  & 83.72   \\ 
        ~ & GatingScore & 60.00\% & 92.6  & 74.0  & 79.8  & \textbf{94.5}  & 93.1  & 86.80   \\ 
        ~ & EASY-EP & 60.00\% & 93.5  & 73.3  & \textbf{81.1}  & \textbf{94.5}  & 93.2  & 87.12   \\ 
        ~ & TENP & 60.00\% & \textbf{93.6} & \textbf{75.1} & \textbf{81.1}  & 94.4  & \textbf{93.3}  & \textbf{87.50}   \\ 
    \bottomrule
    \end{tabular}
}
\caption{The evaluation results of our method on Qwen3-Next-80B-A3B-Instruct at 60\% expert sparsity.}
\label{tab:80BModel}
\end{table*}

\section{Other layer selection methods}
\label{sec:OtherLayerSelectionMethods}
We reproduced both methods on our model using their official open-source implementations. For MoDeGPT (BI), we computed BI scores with the released code and allocated the retained experts to each layer proportionally; when the allocated number exceeded a layer’s capacity, the overflow experts were re-assigned to the layers with the smallest retention. For ShortGPT ranking, we followed the open-source procedure to estimate layer importance and then used the resulting ranking to reorder our trapezoidal per-layer expert allocation. The results indicate that, although layer evaluation criteria developed for dense models can affect expert allocation in MoE models, they do not lead to consistent gains under the same sparsity budget. Specifically, the BI and ShortGPT based variants show slight advantages on language understanding tasks such as ARC-E and ARC-C, but perform noticeably worse on reasoning and code generation benchmarks, including GSM8K, MBPP, and HumanEval. By contrast, our allocation strategy achieves the best overall average performance, while also delivering the strongest results on reasoning and code generation under the same sparsity constraint.

\section{Performance under the 80B parameter setting}
\label{sec:80BModel}
In addition to the two 14B and 16B models used in the main experiments, we also conducted experiments on the Qwen3-Next-80B-A3B-Instruct model. Table~\ref{tab:80BModel} further supplements the validation of the effectiveness of our method on large-scale models. As can be observed from the experimental results, as the number of model parameters increases, the level of redundancy also grows, leading to a smaller loss in model accuracy after pruning. Notably, even random pruning achieves relatively strong performance, with an average score decrease of only about 10\%. In contrast, our methods, TENP and EASY-EP, incur almost no degradation in model accuracy.

\end{document}